\documentclass{article} % For LaTeX2e
\usepackage{iclr2023_conference,times}

\usepackage{amsmath,amsfonts,bm}

\def\eqref#1{equation~\ref{#1}}
\def\Eqref#1{Equation~\ref{#1}}
\def\1{\bm{1}}

\DeclareMathAlphabet{\mathsfit}{\encodingdefault}{\sfdefault}{m}{sl}
\SetMathAlphabet{\mathsfit}{bold}{\encodingdefault}{\sfdefault}{bx}{n}

\newcommand{\E}{\mathbb{E}}

\newcommand{\R}{\mathbb{R}}

\usepackage{hyperref}
\usepackage{url}

\usepackage{amssymb}
\usepackage{amsthm}
\usepackage{mathtools}
\usepackage{empheq}
\usepackage{subcaption}
\usepackage{wrapfig}
\usepackage{graphicx}
\usepackage{xspace}
\usepackage{mathtools}
\usepackage{algorithm}
\usepackage[noend]{algorithmic}
\usepackage{booktabs}
\usepackage{bbm}
\usepackage{hyperref}
\usepackage{url}
\usepackage[compact]{titlesec} 
\hypersetup{colorlinks=true,linkcolor=black,citecolor=brown,urlcolor=blue}
\usepackage{enumitem}

\def\sqrtexplained#1{%
  \begingroup
    \sbox0{$#1$}
    \def\underbrace##1_##2{##1}
    \sbox2{$#1$}
    \dimen0=\wd0 \advance\dimen0-\wd2
    \mathrlap{\sqrt{\phantom{\displaystyle#1}\kern\dimen0 }}
    \hphantom{\sqrt{\vphantom{\displaystyle#1}}}
  \endgroup
  #1}

\def\mdp{\mathcal{M}}

\def\Dset{\mathcal{D}}

\def\Reward{\mathcal{R}}
\def\Trans{\mathcal{T}}

\def\prob{\mathrm{Pr}}

\def\dkl{D_\mathrm{KL}}

\newtheorem{theorem}{Theorem}
\newtheorem{lemma}[theorem]{Lemma}

\newtheorem{assumption}[theorem]{Assumption}
\newtheorem{definition}[theorem]{Definition}

\def\return{R}
\def\RCSL{\mathcal{L}_\mathrm{RCSL}}
\def\VAE{\mathcal{L}_\mathrm{VAE}}
\def\DOC{\mathcal{L}_\mathrm{\method}}

\def\method{DoC\xspace}
\def\vae{VAE\xspace}

\title{Dichotomy of Control: Separating What You Can Control from What You Cannot}
\author{
  Mengjiao Yang \\ University of California, Berkeley \\ Google Research, Brain Team \\ \texttt{sherryy@google.com}
  \And Dale Schuurmans \\ University of Alberta \\ Google Research, Brain Team 
  \And Pieter Abbeel \\ University of California, Berkeley
  \And Ofir Nachum \\ Google Research, Brain Team
}

\iclrfinalcopy % Uncomment for camera-ready version, but NOT for submission.
\begin{document}

\maketitle

\begin{abstract}
Future- or return-conditioned supervised learning is an emerging paradigm for offline reinforcement learning (RL), %in which 
where the future outcome (i.e., return) associated with an observed action sequence %in an offline dataset 
is used as input to a policy trained to imitate those same actions. %as an alternative to learning RL algorithms offline and achieved competitive performance. 
While return-conditioning is at the heart of popular algorithms such as decision transformer (DT), these methods tend to perform poorly in highly stochastic environments, where an occasional high return %associated with an action sequence %might be due more to the randomness of the environment than to the actions themselves. %can lead to overly optimistic policies. 
can arise from randomness in the environment rather than the actions themselves.
Such situations can lead to a learned policy that is \emph{inconsistent} with its conditioning inputs; i.e., using the policy to act in the environment, % --- while 
when %conditioned 
conditioning
on a specific desired return, % --- 
%can lead 
leads
to a distribution of real returns that is wildly different than desired.
%
% [Ofir: I think this is a bit too technical for the abstract] While learning a latent variable of stochastic futures and conditioning the policy on latent futures as opposed to scalar returns seems to be a natural approach to solving this problem, policies obtained through this approach either become too conservative (replicating the data behavior) or still heavily depend on environment stochasticity under which their performance cannot be guaranteed. 
%
In this work, we propose the \emph{dichotomy of control} (DoC), a future-conditioned supervised learning framework that separates mechanisms within a policy's control (actions) from those %outside of 
beyond
a policy’s control (environment stochasticity). 
We achieve this separation by conditioning the policy on a latent variable representation of the future, and designing a mutual information constraint that removes any information from the latent variable associated with randomness in the environment.
Theoretically, we show that \method yields policies that are \emph{consistent} with their conditioning inputs, ensuring that conditioning a learned policy on a desired high-return future outcome will correctly induce high-return behavior. 
Empirically, we show that \method is able to achieve significantly better performance than DT on environments that have highly stochastic rewards %(e.g., Bandit) 
and transitions\footnote{Code available at \url{https://github.com/google-research/google-research/tree/master/dichotomy_of_control}.}. % (e.g., FrozenLake).
\end{abstract}

\section{Introduction}\label{sec:intro}
Offline reinforcement learning (RL) aims to extract an optimal policy solely from an existing %static 
dataset of previous interactions~\citep{fujimoto2019off,wu2019behavior,kumar2020conservative}. As researchers begin to scale offline RL to large image, text, and video datasets~\citep{agarwal2020optimistic,fan2022minedojo,baker2022video,reed2022generalist,reid2022can}, a family of methods known as \emph{return-conditioned supervised learning} (RCSL), including Decision Transformer (DT)~\citep{chen2021decision,lee2022multi} and RL via Supervised Learning (RvS)~\citep{emmons2021rvs}, have gained popularity due to their algorithmic simplicity and ease of scaling. 
At the heart of RCSL is the idea of conditioning a policy on a specific future outcome, often a return~\citep{srivastava2019training,kumar2019reward,chen2021decision} but also sometimes a goal state or generic future event~\citep{codevilla2018end,ghosh2019learning,lynch2020learning}. RCSL trains a policy to imitate actions associated with a conditioning input via supervised learning. During inference (i.e., at evaluation), the policy is conditioned on a desirable high-return or future outcome, with the hope of inducing behavior that can achieve this desirable outcome.

\begin{figure}[t]
    \centering
    \includegraphics[width=.8\textwidth]{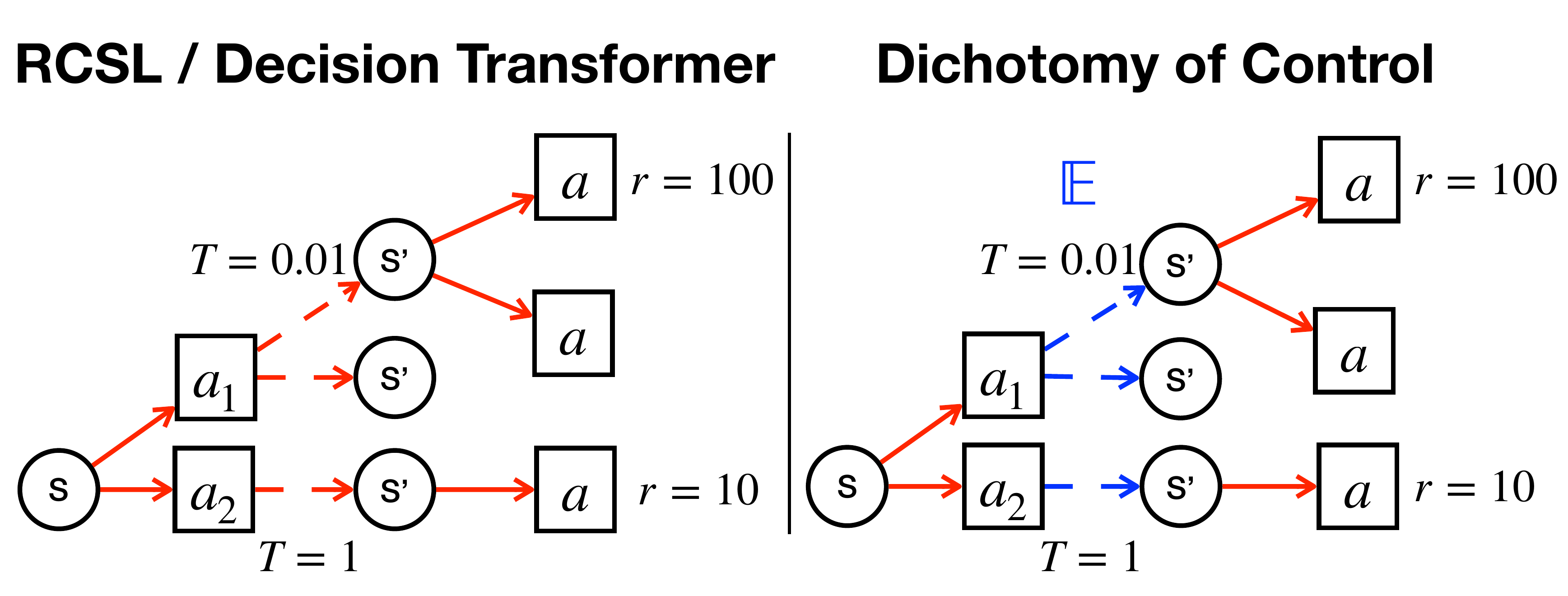}
    \caption{Illustration of DT (RCSL) and \method. Circles and squares denote states and actions. % in an MDP. 
    Solid arrows denote policy decisions. Dotted arrows denote (stochastic) environment transitions. All arrows and nodes are present in the dataset, i.e., there are 4 %(types of)
    trajectories, 2 of which achieve 0 reward. DT {\color{red}maximizes} returns across an entire trajectory, leading to suboptimal policies when a large return ($r=100$) is achieved only due to very low-probability environment transitions ($T=0.01$). \method separates policy stochasticity from that of the environment and only tries to control action decisions (solid arrows), achieving optimal control through maximizing {\color{blue}expected} returns at each timestep.}
    \label{fig:toy}
\end{figure}

Despite the empirical advantages that come with supervised training~\citep{emmons2021rvs,kumar2021should}, RCSL can be highly suboptimal in stochastic environments~\citep{brandfonbrener2022does}, where the future an RCSL policy conditions on (e.g., return) can be primarily determined by randomness in the environment rather than the data collecting policy itself. Figure~\ref{fig:toy} (left) illustrates an example, where conditioning an RCSL policy on the highest return observed in the dataset ($r = 100$) leads to a policy ($a_1$) that relies on a stochastic transition of very low probability ($T = 0.01$) to achieve the desired return of $r=100$; by comparison the choice of $a_2$ is much better in terms of average return, as it surely achieves $r = 10$. 
The crux of the issue %highlighted here 
is that the RCSL policy is \emph{inconsistent} with its conditioning input. Conditioning the policy on a desired return (i.e., $100$) to act in the environment leads to a distribution of real returns (i.e., $0.01*100$) that is wildly different from the return value being conditioned on.
This issue would not have occurred if the %RCSL 
policy could also maximize the transition probability that led to the high-return state, but this is not possible as transition probabilities are a part of the environment and not subject to the policy’s control.

%\sherry{Do we want to remove this paragraph?} A na\"ive workaround to the aforementioned inconsistency issue is to condition an RCSL policy on the \emph{expected} returns of \emph{immediate} actions. This corresponds to RL methods that fit the state or state-action value function of the \emph{behavior} policy that collected the dataset~\citep{peng2019advantage,brandfonbrener2021offline}.
%However, this leads to overly conservative policies that only do a ``one-step'' improvement (i.e., the first maximum over actions), which is suboptimal compared to taking the best action at each step as shown in Figure~\ref{fig:toy} (right).

%One natural approach that might come to mind as a solution to the above example is to somehow average the returns or cluster the trajectories in the training data before using the average returns (clusters) as conditioning inputs when learning the policy. 
A number of works propose a generalization of RCSL, known as future-conditioned supervised learning methods. These techniques have been shown to be effective in imitation learning~\citep{singh2020parrot,pertsch2020accelerating}, offline Q-learning~\citep{ajay2020opal}, and online policy gradient~\citep{venuto2021policy}. 
It is common in future-conditioned supervised learning to apply a KL divergence regularizer  on the latent variable -- inspired by variational auto-encoders (VAE)~\citep{kingma2013auto} and measured with respect to a learned prior conditioned only on past information -- to limit the amount of future information captured in the latent variable. %, often motivated as a form of regularization.
It is natural to ask whether this regularizer could remedy the insconsistency of RCSL. %future-conditioned supervised learning could remedy the inconsistency issue of RCSL by learning a latent variable representation of the future and applying a KL regularizer to limit the amount of future information the latent variable captures. 
Unfortunately, as the KL regularizer makes no distinction between future information that is controllable versus that which is not, such an % na\"ive 
approach will still exhibit inconsistency, in the sense that the latent variable representation may contain information about the future that is due only to environment stochasticity. % if the regularizer is too weak
%if the regularizer is too strong, the objective may reduce to behavioral cloning, limiting the ability of the learning algorithm to learn a policy better than the data collecting policy~\citep{ajay2020opal}. 
%, or suffer from being overly conservative if the regularizer is too strong.

It is clear that the major issue with both RCSL and na\"ive variational methods is that they make no distinction between stochasticity of the policy (controllable) and stochasticity of the environment (uncontrollable). An optimal policy should maximize over the controllable (actions) and take expectations over uncontrollable (e.g., transitions) as shown in Figure~\ref{fig:toy} (right). This implies that, under a variational approach, the latent variable representation %of the future 
that a policy conditions on should not incorporate any information that is solely due to randomness in the environment. In other words, while the latent representation can and should include information about future behavior (i.e., actions), it should not reveal any information about the rewards or transitions associated with this behavior.

To this end, we propose a future-conditioned supervised learning framework termed \emph{dichotomy of control} (\method), which, in Stoic terms~\citep{shapiro2014wrote}, has \emph{``the serenity to accept the things it cannot change, courage to change the things it can, and wisdom to know the difference.''} \method separates mechanisms within a policy’s control (actions) from those beyond a policy’s control (environment stochasticity). To achieve this separation, we condition the policy on a latent variable representation of the future while minimizing the mutual information between the latent variable and future stochastic rewards and transitions in the environment. By only capturing the controllable factors in the latent variable, \method can maximize over each action step without also attempting to maximize environment transitions as shown in Figure~\ref{fig:toy} (right). Theoretically, we show that \method policies are \emph{consistent} with their conditioning inputs, ensuring that conditioning on a high-return future will correctly induce high-return behavior. Empirically, we show that \method can outperform both RCSL and na\"ive variational methods on highly stochastic %Bandit and FrozenLake
environments.

\section{Related Work}
\paragraph{Return-Conditioned Supervised Learning.} Since offline RL algorithms~\citep{fujimoto2019off,wu2019behavior,kumar2020conservative} can be sensitive to hyper-parameters and difficult to apply in practice~\citep{emmons2021rvs,kumar2021should}, return-conditioned supervised learning (RCSL) has become a popular alternative, particularly when the environment is deterministic and near-expert demonstrations are available~\citep{brandfonbrener2022does}. RCSL learns to predict behaviors (actions) by conditioning on desired returns~\citep{schmidhuber2019reinforcement,kumar2019reward} using an MLP policy~\citep{emmons2021rvs} or a transformer-based policy that encapsulates history~\citep{chen2021decision}. Richer information other than returns, such as goals~\citep{codevilla2018end,ghosh2019learning} or trajectory-level aggregates~\citep{furuta2021generalized}, have also been used as inputs to a conditional policy in practice. Our work also conditions policies on richer trajectory-level information in the form of a latent variable representation of the future, with additional theoretical justifications of such conditioning in stochastic environments.

\paragraph{RCSL Failures in Stochastic Environments.} Despite the empirical success of RCSL achieved by DT and RvS, recent work has noted the failure modes in stochastic environments. \citet{paster2020planning} and \citet{vstrupl2022upside} presented counter-examples where online RvS can diverge in stochastic environments. \citet{brandfonbrener2022does} identified near-determinism as a necessary condition for RCSL to achieve optimality guarantees similar to other offline RL algorithms but did not propose a solution for RCSL in stochastic settings. \citet{paster2022you} identified this same issue with stochastic transitions and proposed to cluster offline trajectories and condition the policy on the average cluster returns. However, the approach in~\citet{paster2022you} %(1) 
has technical limitations %theoretical flaws 
(see Appendix~\ref{sec:bad-consistency}), %(2) 
does not account for reward stochasticity, and %(3) 
still conditions the policy on (expected) returns, which can lead to undesirable policy-averaging, i.e., a single policy covering two very different behaviors (clusters) that happen to have the same return. \citet{villaflor2022addressing} also identifies overly optimistic behavior of DT and proposes to use discrete $\beta$-VAE to induce diverse future predictions a policy can condition on. This approach only differs the issue with stochastic environments to stochastic latent variables, i.e., the latent variables will still contain stochastic environment information that the policy cannot reliably reproduce.

%The first approach that might come to mind is to somehow average the returns observed in the training data before using them as conditioning inputs when learning $\pi$. For example, one could first cluster the training data according to some criterion before averaging the returns in each cluster and using these cluster averages to condition a policy during RCSL training~\citep{paster2022you}. Even if one assumes the clusters are devised correctly (to avoid both overly large clusters which lead to conservative policies like in one-step RL, or overly small clusters which lead to inconsistency), conditioning on expected return can still lead to undesirable policy-averaging. If two clusters have identical average returns, then return-conditioned learning of these clusters will train a single policy to cover two potentially very different behaviors.

\paragraph{Learning Latent Variables from Offline Data.} Various works have explored learning a latent variable representation of the future (or past) transitions in offline data via maximum likelihood and use the latent variable to assist planning~\citep{lynch2020learning}, imitation learning~\citep{kipf2019compile,ajay2020opal,hakhamaneshi2021hierarchical}, offline RL~\citep{ajay2020opal,zhou2020plas}, or online RL~\citep{fox2017multi,krishnan2017ddco,goyal2019recurrent,shankar2020learning,singh2020parrot,wang2021skill,venuto2021policy}. These works generally focus on the benefit of increased temporal abstraction afforded by using latent variables as higher-level actions in a hierarchical policy. \citet{villaflor2022addressing} has introduced latent variable models into RCSL, which is one of the essential tools that enables our method, but they did not incoporate the appropriate constraints which can allow RCSL to effectively combat environment stochasticity, as we will see in our work.
\section{Preliminaries}

\paragraph{Environment Notation}
We consider the problem of learning a decision-making agent to interact with a sequential, finite-horizon environment. At time $t=0$, the agent observes an initial state $s_0$ determined by the environment. After observing $s_t$ at a timestep $0\le t\le H$, the agent chooses an action $a_t$. After the action is applied %to the environment and 
the environment yields an immediate scalar reward $r_t$ and, if $t<H$, a next state $s_{t+1}$. We use $\tau:=(s_t,a_t,r_t)_{t=0}^H$ to denote a generic \emph{episode} generated from interactions with the environment, and use $\tau_{i:j}:=(s_t,a_t,r_t)_{t=i}^j$ to denote a generic \emph{sub-episode}, with the understanding that $\tau_{0:-1}$ refers to an empty sub-episode. The \emph{return} associated with an episode $\tau$ is defined as $\return(\tau):= \sum_{t=0}^H r_t$. 

We will use $\mdp$ to denote the environment. We assume that $\mdp$ is determined by a stochastic reward function $\Reward$, stochastic transition function $\Trans$, and unique initial state $s_0$, so that $r_t\sim \Reward(\tau_{0:t-1}, s_t, a_t)$ and $s_{t+1}\sim \Trans(\tau_{0:t-1}, s_t, a_t)$ during interactions with the environment. Note that these definitions specify a history-dependent environment, as opposed to a less general Markovian environment.

\paragraph{Learning a Policy in RCSL}
In future- or return-conditioned supervised learning, one uses a fixed training data distribution $\Dset$ of episodes $\tau$ (collected by unknown and potentially multiple agents) to learn a policy $\pi$,
where $\pi$ is trained to predict $a_t$ conditioned on the history $\tau_{0:t-1}$, the observation $s_t$, and an additional conditioning variable $z$ that may depend on both the past and future of the episode. For example, in return-conditioned supervised learning, policy training minimizes the following objective over $\pi$:
\begin{equation}
    \label{eq:rcsl-obj}
    \RCSL(\pi) := \E_{\tau\sim\Dset}\left[\sum_{t=0}^H -\log \pi (a_t|\tau_{0:t-1},s_t,z(\tau)) \right],
\end{equation}
where $z(\tau)$ is the return $\return(\tau)$.

\paragraph{Inconsistency of RCSL}
To apply an RCSL-trained policy $\pi$ during inference --- i.e., interacting online with the environment --- one must first choose a specific $z$.\footnote{For simplicitly, we assume $z$ is chosen at timestep $t=0$ and held constant throughout an entire episode. As noted in~\citet{brandfonbrener2022does}, this protocol also encompasses instances like DT~\citep{chen2021decision} in which $z$ at timestep $t$ is the (desired) return summed starting at $t$.} For example, one might set $z$ to be the maximal return observed in the dataset, in the hopes of inducing a behavior policy which achieves this high return. Using $\pi_z$ as a shorthand to denote the policy $\pi$ conditioned on a specific $z$, we define the expected return $V_{\mdp}(\pi_z)$ of $\pi_z$ in $\mdp$ as, 
\begin{equation}
    V_{\mdp}(\pi_z) := \E_{\tau\sim \prob[\cdot|\pi_z,\mdp]}\left[R(\tau)\right].
\end{equation}
Ideally the expected return induced by $\pi_z$ is close to $z$, i.e., $z\approx V_{\mdp}(\pi_z)$, so that acting according to $\pi$ conditioned on a high return induces behavior which actually achieves a high return. However, RCSL training according to~\Eqref{eq:rcsl-obj} will generally yield policies that are highly \emph{inconsistent} in stochastic environments, meaning that the achieved returns may be significantly different than $z$ (i.e., $V_{\mdp}(\pi_z)\neq z$). This has been highlighted in various previous works~\citep{brandfonbrener2022does,paster2022you,vstrupl2022upside,eysenbach2022imitating,villaflor2022addressing}, and we provided our own example in Figure~\ref{fig:toy}.

\paragraph{Approaches to Mitigating Inconsistency} A number of future-conditioned supervised learning approaches propose to learn a stochastic latent variable embedding of the future, $q(z|\tau)$, while regularizing $q$ with a KL-divergence from a learnable \emph{prior} conditioned only on the past $p(z|s_0)$~\citep{ajay2020opal,venuto2021policy,lynch2020learning}, thereby minimizing:
\begin{align}
    \label{eq:vae}
    \VAE(\pi, q, p) := \E_{\tau\sim\Dset,z\sim q(z|\tau)}\left[\sum_{t=0}^H -\log \pi (a_t|\tau_{0:t-1},s_t,z) \right]+\beta\cdot\E_{\tau\sim\Dset}\left[\dkl(q(z|\tau)\|p(z|s_0))\right].
\end{align}
One could consider adopting such a future-conditioned objective in RCSL. However, since the KL regularizer makes no distinction between observations the agent can control (actions) from those it cannot (environment stochasticity), the choice of coefficient $\beta$ applied to the regularizer introduces a `lose-lose' trade-off. Namely, as noted in~\citet{ajay2020opal}, if the regularization coefficient is too large ($\beta\ge1$), the policy will not learn diverse behavior (since the KL limits how much information of the future actions is contained in $z$); while if the coefficient is too small ($\beta<1$), the policy's learned behavior will be inconsistent (in the sense that $z$ will contain information of environment stochasticity that the policy cannot reliably reproduce). The discrete $\beta$-VAE incoporated by \citet{villaflor2022addressing} with $\beta < 1$ corresponds to this second failure mode.
\section{Dichotomy of Control}\label{sec:method}
In this section, we first propose the \method objective for learning future-conditioned policies that are guaranteed to be consistent. We then present a practical framework for optimizing \method's constrained objective in practice and an inference scheme to enable better-than-dataset behavior via a learned value function and prior. 

\subsection{Dichotomy of Control via Mutual Information Minimization}
As elaborated in the prevous section, whether $z(\tau)$ is the return $\return(\tau)$ or more generally a stochastic latent variable with distribution $q(z|\tau)$, existing RCSL methods fail to satisfy consistency because they insufficiently enforce the type of future information $z$ can contain. %The major caveat in the RCSL objective in~\Eqref{eq:rcsl-obj} that led to its inconsistency is that the policy conditions on $z(\tau) = R(\tau)$, where $\tau$ depends on the stochasticity of the environment that a policy has no control over. What \emph{is} under a policy's control, in fact, is some future statistic $q(\tau)$, that distills $\tau$ into a latent variable $z$, where $z$ does not contain any information due to the randomness of the environment. 
Our key observation is that $z$ should not include any information due to environment stochasticity, i.e., any information about a future $r_t,s_{t+1}$ that is not already known given the previous history up to that point $\tau_{0:t-1},s_t,a_t$.
Accordingly, we modify the RCSL objective from Equation~\ref{eq:rcsl-obj} to incorporate a conditional mutual information constraint between $z$ and each pair $r_t,s_{t+1}$ in the future:
\begin{align}
    \DOC(\pi,q) := & ~\E_{\tau\sim\Dset,z\sim q(z|\tau)}\left[\sum_{t=0}^H -\log \pi (a_t|\tau_{0:t-1},s_t,z)\right]\nonumber\\
    \mathrm{s.t.}~~~ & \mathrm{MI}(r_t; z ~|~\tau_{0:t-1},s_t,a_t) = 0, \mathrm{MI}(s_{t+1}; z ~|~\tau_{0:t-1},s_t,a_t) = 0, \label{eq:doc-constrained} \\
    &~\forall~\tau_{0:t-1},s_t,a_t ~\text{and}~ 0\leq t \leq H,
\end{align}
where $\mathrm{MI}(r_t; z |\tau_{0:t-1},s_t,a_t)$ denotes the mutual information between $r_t$ and $z$ given $\tau_{0:t-1}, s_t, a_t$ when measured under samples of $r_t,z$ from $\Dset,q$;
and analogously for $\mathrm{MI}(s_{t+1}; z |\tau_{0:t-1},s_t,a_t)$. 
%The meaning of $\mathrm{MI}(s_{t+1}; z |\tau_{0:t-1},s_t,a_t)$ is analogous.

The first part of the \method objective conditions the policy on a latent variable representation of the future, similar to the first part of the future-conditioned VAE objective in~\Eqref{eq:vae}. However, unlike~\Eqref{eq:vae}, the \method objective enforces a much more precise constraint on $q$, given by the MI constraints in~\Eqref{eq:doc-constrained}. % will ensure that $z\sim q(\tau)$ does not capture any randomness from $r$ and $s'$.

\subsection{Dichotomy of Control in Practice}
\paragraph{Contrastive Learning of \method Constraints.} To satisfy the mutual information constraints in~\Eqref{eq:doc-constrained} we transform the MI to a contrastive learning objective. Specifically, for the constraint on $r$ and $z$ (and similarly on $s_{t+1}$ and $z$) one can derive,
\begin{align}
\hspace{-1mm}\mathrm{MI}(&r_t; z |\tau_{0:t-1},s_t,a_t)\nonumber\\=&~\dkl\left(\prob[r_t , z|\tau_{0:t-1},s_t,a_t]\|\prob[r_t|\tau_{0:t-1},s_t,a_t]\prob[z|\tau_{0:t-1},s_t,a_t]\right)\nonumber\\
=&~\E_{\prob[r_t,z|\tau_{0:t-1},s_t,a_t]}\left[\log\left(\frac{\prob[r_t|z,\tau_{0:t-1},s_t,a_t]}{\prob[r_t|\tau_{0:t-1},s_t,a_t]}\right)\right]\nonumber\\
=&~\E_{\prob[r_t,z|\tau_{0:t-1},s_t,a_t]}\log\prob[r_t|z,\tau_{0:t-1},s_t,a_t] - \E_{\prob[r_t|\tau_{0:t-1},s_t,a_t]}\log\prob[r_t|\tau_{0:t-1},s_t,a_t].\label{eq:mi-kl}
\end{align}
The second expectation above is a constant with respect to $z$ and so can be ignored during learning. We further introduce a conditional distribution $\omega(r_t|\tau_{0:t-1},s_t,a_t)$ parametrized by an energy-based function $f:\Omega\mapsto\R$: 
\begin{equation}
\omega(r_t|z,\tau_{0:t-1},s_t,a_t) \propto \rho(r_t)\exp{\{f(r_t, z, \tau_{0:t-1},s_t,a_t)\}},
\end{equation}
where $\rho$ is some fixed sampling distribution of rewards. In practice, we set $\rho$ to be the marginal distribution of rewards in the dataset.
Hence we express the first term of Equation~\ref{eq:mi-kl} via an optimization over $\omega$, i.e.,
\begin{align*}
&\max_\omega~\E_{\prob[r_t,z|\tau_{0:t-1},s_t,a_t]}\left[\log\omega(r_t|\tau_{0:t-1},s_t,a_t)\right]\nonumber\\
=&\max_f\E_{\prob[r_t,z|\tau_{0:t-1},s_t,a_t]}\left[f(r_t, z, \tau_{0:t-1},s_t,a_t) - \log \E_{\rho(\tilde{r})}\left[\exp \{f(\tilde{r}, z, \tau_{0:t-1},s_t,a_t)\}\right]\right].
\end{align*}
Combining this (together with the analogous derivation for $\mathrm{MI}(s_{t+1}; z |\tau_{0:t-1},s_t,a_t)$) with Equation~\ref{eq:doc-constrained} via the Lagrangian, we can learn $\pi$ and $q(z|\tau)$ by minimizing the final \method objective:
%\ofir{In this objective you have an expectation over $z$ outside the log, but in the expression above you have the expectation over $z$ inside the log. I think both are `fine' in terms of minimizing MI, but the derivation does not match the result.} 
%\sherry{I updated the equation above to use expectation outside the log, but I am confused about the Lagrangian derivation below. With the above supremum being $\leq 0$, shouldn't we also be optimizing the Lagrangian multipliers $\beta$ below?}
\begin{align}
    &\DOC(\pi,q) = \max_{f}\E_{\tau\sim\Dset,z\sim q(z|\tau)}\left[\sum_{t=0}^H -\log \pi (a_t|\tau_{0:t-1},s_t,z) \right]\nonumber\\
    +&\beta\cdot\sum_{t=0}^{H}\E_{\tau\sim\Dset,z\sim q(z|\tau)}\left[f(r_t, s_{t+1}, z, \tau_{0:t-1},s_t,a_t) - \log \E_{\rho(\tilde{r},\tilde{s}')}\left[\exp \{f(\tilde{r}, \tilde{s}', z, \tau_{0:t-1},s_t,a_t)\}\right]\right].\label{eq:doc}
\end{align}

%\begin{wrapfigure}{r}{\textwidth}
\begin{minipage}{\linewidth}
\begin{algorithm}[H]
\caption{Inference with Dichotomy of Control}
\begin{algorithmic}
\STATE \textbf{Inputs}~Policy $\pi(\cdot|\cdot, \cdot, \cdot)$, prior $p(\cdot)$, value function $V(\cdot)$, 
%interaction history $\tau_{0:t-1}$, 
initial state $s_0$, %current state $s_t$, 
number of samples hyperparameter $K$.
\STATE Initialize $z^*; V^*$ \hfill\texttt{\footnotesize $\rhd$ Track the best latent and its value.}
    \FOR {$k = 1$ to $K$}
    	\STATE Sample $z_k\sim p(z|s_0)$ \hfill\texttt{\footnotesize $\rhd$ Sample a latent from the learned prior.}
        \IF {$V(z_k) > V^*$} 
    	\STATE $z^* = z_k$; $V^* = V$ \hfill\texttt{\footnotesize $\rhd$ Set best latent to the one with the highest value.}
        \ENDIF
    \ENDFOR\,\textbf{return} $\pi(\cdot|\cdot,\cdot, z^*)$ \hfill\texttt{\footnotesize $\rhd$ Policy conditioned on the best $z^*$.}
\end{algorithmic}
\label{algo:doc-inference}
\end{algorithm}
\end{minipage}
%\end{wrapfigure}

\paragraph{\method Inference.}
As is standard in RCSL approaches, the policy learned by \method requires an appropriate conditioning input $z$ to be chosen during inference.
To choose a desirable $z$ associated with high return, we propose to (1) enumerate or sample a large number of potential values of $z$, (2) estimate the expected return for each of these values of $z$, (3) choose the $z$ with the highest associated expected return to feed into the policy. 
To enable such an inference-time procedure, we need to add two more components to the method formulation: %\method objective. 
First, a prior distribution $p(z|s_0)$ from which we will sample a large number of values of $z$; second, a value function $V(z)$ with which we will rank the potential values of $z$. These components are learned by minimizing the following objective:
%\ofir{The motivation above suggests that we should stopgrad $z$ when input to $V$ as well, but I actually don't think that this would be a good thing algorithmically...}
%\sherry{It was not obvious to me why the above motivation suggests we need a stopgrad on $z$ when inputting to $V$. The motivation for stopgrad on $z$ inputting to the prior's KL below is so that $q$ is not regularized by the prior, which is also learned, but why is it not okay to have $V$'s gradient affect $q$?}
\begin{equation}\label{eq:doc-aux}
    \mathcal{L}_\text{aux}(V, p) = \E_{\tau\sim\Dset,z\sim q(z|\tau)}\bigg[\left(V(z) - \return(\tau)\right)^2 %\nonumber\\&
    +\dkl(\text{stopgrad}(q(z|\tau))\|p(z|s_0))\bigg].
\end{equation}
Note that we apply a stop-gradient to $q(z|\tau)$ when learning $p$ so as to avoid regularizing $q$ with the prior. This is unlike the VAE approach, which by contrast advocates \emph{for} regularizing $q$ via the prior.  
%During inference, a set of future statistics $(z_1,...z_K)$ are sampled according to $z_k\sim p(s_0)$ and ranked according to $V(z_k)$. The policy chooses the best future statistic, $z^* = \argmax_{z_k} V(z_k)$, and conditions on this statistic to act in the environment, i.e., $\pi(a_t|\tau_{0:t-1}, s_t, z^*)$. 
See Algorithm~\ref{algo:doc-inference} for inference pseudocode %of running \method during inference 
(and Appendix~\ref{app:pseudo} for training pseudocode). % of training).
\section{Consistency Guarantees for Dichotomy of Control}\label{sec:analysis}

We %use this section to theoretical analyze our
provide a theoretical justification of the
proposed learning objectives $\DOC$ and $\mathcal{L}_\text{aux}$,
showing that, if they are minimized, the resulting
%that demonstrates the resulting
%Namely, we provide a guarantee that our 
inference-time procedure %is 
will be
sound, in the sense that \method will learn a $V$ and $\pi$ such that the true value of $\pi_z$ in the environment $\mdp$ is equal to $V(z)$. 
More specifically we define the following notion of \emph{consistency}:
%Since \method relies on a value function $V(z)$ to act in an environment as discussed in the previous section, we need to ensure that $\pi$ and $V$ are consistent for some $z$ (i.e., acting according
%to $\pi$ conditioned on some $z$ induces behavior that actually achieves $V(z)$ in the environment). We present the necessary assumptions and \method's consistency guarantees in this section.

\begin{definition}[Consistency]
\label{def:consistency}
A future-conditioned policy $\pi$ and value function $V$ are \textbf{consistent} for a specific conditioning input $z$ if the expected return of $z$ predicted by $V$ is equal to the true expected return of $\pi_z$ in the environment: $V(z) = V_{\mdp}(\pi_z)$. %Future-conditioned policy $\pi$ and value distribution $V$ are $\epsilon-$\textbf{consistent} for a specific conditioning input $z$ if $\dtv(V(z) \| V_{\mdp}(\pi_z)) \le \epsilon$. 
\end{definition}

To guarantee consistency of $\pi,V$, we will make the following two assumptions: 
%First, an assumption ensuring that the training dataset $\Dset$ provides data collected from the environment $\mdp$; second, a simplifying assumption on \method that allows us to ignore any optimization and approximation errors associated with the objective in practice.

\begin{assumption}[Data and environment agreement] 
\label{ass:data-env}
The per-step reward and next-state transitions observed in the data distribution are the same as those of the environment. In other words, for any $\tau_{0:t-1},s_t,a_t$ with $\prob[\tau_{0:t-1},s_t,a_t|\Dset] > 0$, we have $\prob[\hat{r}_t=r_t|\tau_{0:t-1},s_t,a_t,\Dset]=\Reward(\hat{r}_t|\tau_{0:t-1},s_t,a_t)$ and $\prob[\hat{s}_{t+1}=s_{t+1}|\tau_{0:t-1},s_t,a_t,\Dset]=\Trans(\hat{s}_{t+1}|\tau_{0:t-1},s_t,a_t)$ for all $\hat{r}_t,\hat{s}_{t+1}$.
\end{assumption}

\begin{assumption}[No optimization or approximation errors] \label{ass:bayes}
\method yields policy $\pi$ and value function $V$ that are Bayes-optimal with respect to the training data distribution and $q$. In other words, $V(z) = \E_{\tau\sim\prob[\cdot | z, \Dset]}\left[\return(\tau)\right]$ and $\pi(\hat{a} | \tau_{0:t-1}, s_t, z) = \prob\left[\hat{a} = a_t | \tau_{0:t-1}, s_t, z, \Dset\right]$.
\end{assumption}

%With these two assumptions in hand, 
Given these two assumptions,
we can then establish the following %guarantee on the consistency of $V,\pi$ given by \method.
consistency guarantee for \method.
\begin{theorem}
\label{thm:consistency}
Suppose \method yields $\pi, V, q$ with $q$ satisfying the MI constraints:
\begin{equation}
    \mathrm{MI}(r_t;z|\tau_{0:t-1},s_t,a_t)=\mathrm{MI}(s_{t+1};z|\tau_{0:t-1},s_t,a_t)=0,
\end{equation}
for all $\tau_{0:t-1},s_t,a_t$ with $\prob[\tau_{0:t-1},s_t,a_t|\Dset] > 0$.
Then under Assumptions~\ref{ass:data-env} and~\ref{ass:bayes}, $V$ and $\pi$ are consistent for any $z$ with $\prob[z|q,\Dset] >0$.
\end{theorem}
For proof, see Appendix~\ref{sec:proof}.

%\paragraph{Remark.} In Appendix~\ref{sec:markov-policies} we show that Theorem~\ref{thm:consistency} can hold even for Markov policies in Markov environments, while alternative notions of consistency are not as generally applicable; see further discussion in Appendix~\ref{sec:bad-consistency}. 

\paragraph{Consistency in Markovian environments.}
While the results above are focused on environments and policies that are non-Markovian, one can extend Theorem~\ref{thm:consistency} to Markovian environments and policies. This result is somewhat surprising, as the assignments of $z$ to episodes $\tau$ induced by $q$ are necessarily history-dependent, and projecting the actions appearing in these clusters to a non-history-dependent policy would seemingly lose important information.
However, %it turns out that 
a Markovian assumption on the rewards and transitions of the environment is sufficient to ensure that no `important' information will be lost, at least in terms of the satisfying requirements for consistency in Definition~\ref{def:consistency}. Alternative notions of consistency are not as generally applicable; see %further discussion in 
Appendix~\ref{sec:bad-consistency}.

We begin by stating our assumptions. %, first the Markov restrictions on the environment $\mdp$ and then an analogue to Assumption~\ref{ass:bayes} in the case that $\pi$ is Markov.
\begin{assumption}[Markov environment]
\label{ass:markov}
The rewards and transitions of $\mdp$ are Markovian; i.e., $\Reward(\tau_{0:t-1},s_t,a_t) = \Reward(\tilde{\tau}_{0:t-1},s_t,a_t)$ and $\Trans(\tau_{0:t-1},s_t,a_t) = \Trans(\tilde{\tau}_{0:t-1},s_t,a_t)$ for all $\tau,\tilde{\tau},s_t,a_t$. We use the shorthand $\Reward(s_t,a_t),\Trans(s_t,a_t)$ for these history-independent functions.
\end{assumption}

\begin{assumption}[Markov policy, without optimization or approximation errors] \label{ass:bayes2}
The policy learned by \method is Markov. This policy $\pi$ as well as its corresponding learned value function $V$ are Bayes-optimal with respect to the training data distribution and $q$. In other words, $V(z) = \E_{\tau\sim\prob[\cdot | z, \Dset]}\left[\return(\tau)\right]$ and $\pi(\hat{a} | s_t, z) = \prob\left[\hat{a} = a_t | s_t, z, \Dset\right]$.
\end{assumption}

With these two assumptions, we can then establish the analogue to Theorem~\ref{thm:consistency}, which relaxes the dependency on history for both the policy $\pi$ and the MI constraints:
\begin{theorem}
\label{thm:consistency2}
Suppose \method yields $\pi, V, q$ with $q$ satisfying the MI constraints:
\begin{equation}
    \mathrm{MI}(r_t;z|s_t,a_t)=\mathrm{MI}(s_{t+1};z|s_t,a_t)=0,
\end{equation}
for all $s_t,a_t$ with $\prob[s_t,a_t|\Dset] > 0$.
Then under Assumptions~\ref{ass:data-env},~\ref{ass:markov}, and~\ref{ass:bayes2}, $V$ and $\pi$ are consistent for any $z$ with $\prob[z|q,\Dset] >0$.
\end{theorem}
For proof, see Appendix~\ref{sec:markov-policies}.
\section{Experiments}\label{sec:exp}

%We now empirically evaluate the effectiveness of \method. We begin our evaluation on a Bernoulli bandit problem with stochastic rewards, based on a canonical `worse-case scenario' for RCSL from~\citet{brandfonbrener2022does}. We then consider the FrozenLake domain from~\citep{brockman2016openai}, where the future VAE approach is ineffective. Finally, we modify a set of OpenAI Gym~\citep{brockman2016openai} environments to introduce high environment stochasticity. \method exhibits significant advantage over RCSL / DT, and outperforms future VAE when the analogous to ``one-step'' RL is insufficient.  For DT, we use the same implementation and hyperparameters as \citet{chen2021decision}. Both \vae and \method builds on the DT implementation and additionally learn a Gaussian latent variable over the next 20 future steps.
We conducted an empirical evaluation to ascertain the effectiveness of \method.  For this evaluation, we considered three settings: (1) a Bernoulli bandit problem with stochastic rewards, based on a canonical `worst-case scenario' for RCSL~\citep{brandfonbrener2022does};  (2) the FrozenLake domain from~\citep{brockman2016openai}, where the future VAE approach proves ineffective; and finally (3) a modified set of OpenAI Gym~\citep{brockman2016openai} environments where we introduced environment stochasticity.  In these studies, we found that \method exhibits a significant advantage over RCSL/DT, and outperforms future VAE when the analogous to ``one-step'' RL is insufficient.  For DT, we use the same implementation and hyperparameters as \citet{chen2021decision}.  Both \vae and \method are built upon the DT implementation and additionally learn a Gaussian latent variable over succeeding 20 future steps.
%, and the coefficient $\beta$ is set to 1 for both methods \ofir{Why not $<1$ for VAE?}. 
See experiment details in Appendix~\ref{app:exp_details} and additional results in Appendix~\ref{app:exp_results}. 

\subsection{Evaluating Stochastic Rewards in Bernoulli Bandit}
\begin{wrapfigure}{r}{0.55\textwidth}
    \centering
    \includegraphics[width=\linewidth]{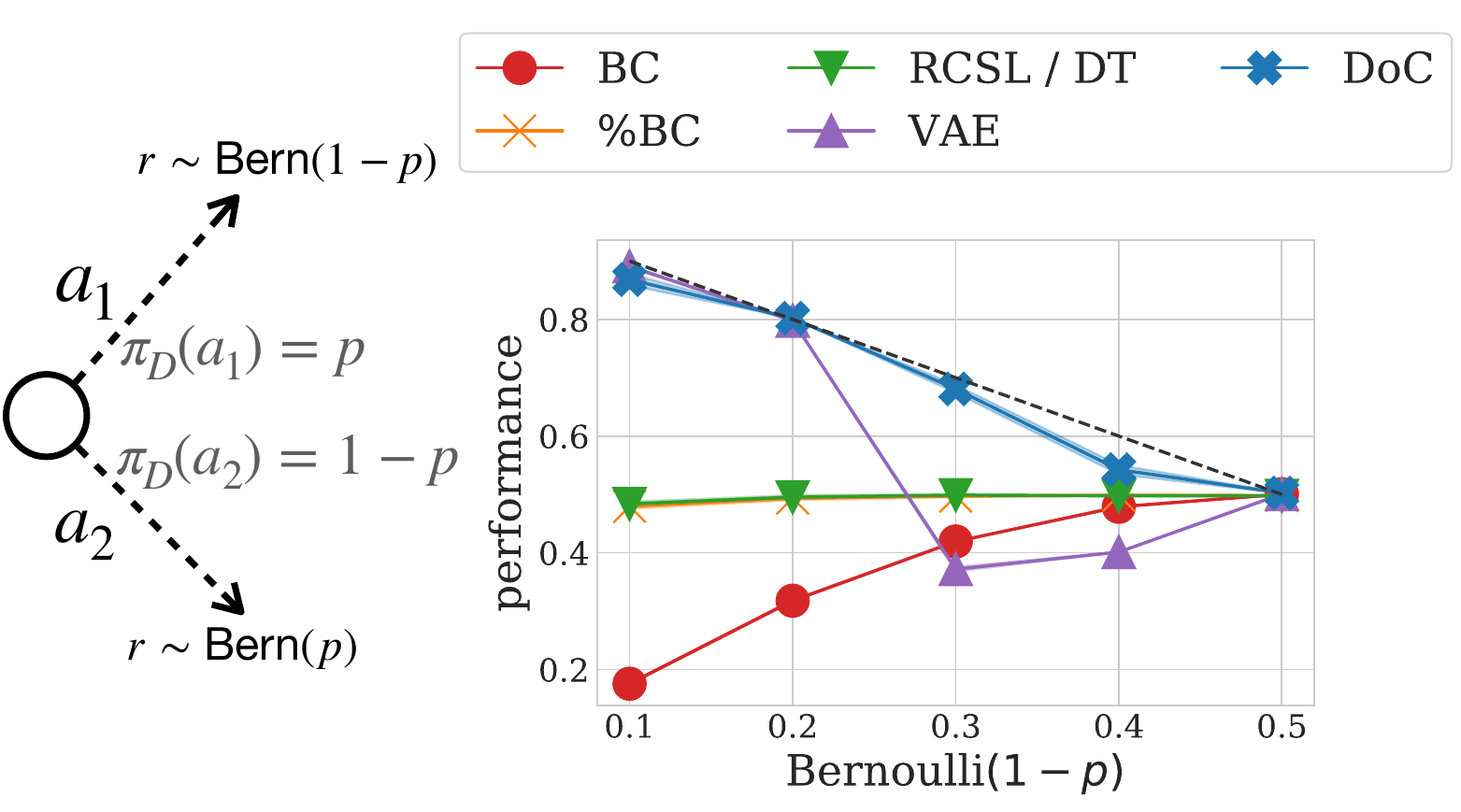}
    \caption{[Left] Bernoulli bandit where the better arm $a_1$ with reward $\texttt{Bern}(1-p)$ for $p<0.5$ is pulled with probability $\pi_D(a_1)=p$ in the offline data. [Right] Average rewards achieved by \method and baselines across 5 environment seeds. RCSL is highly suboptimal when $p$ is small, whereas \method achieves close to Bayes-optimal performance (dotted line) for all values of $p$.}
    \label{fig:bandit}
    \vspace{-0.5cm}
\end{wrapfigure}

\paragraph{Bernoulli Bandit.} Consider a two-armed bandit as shown in Figure~\ref{fig:bandit} (left). The two arms, $a_1, a_2$, have stochastic rewards drawn from Bernoulli distributions of $\texttt{Bern}(1-p)$ and $\texttt{Bern}(p)$, respectively. In the offline dataset, the $a_1$ arm with reward $\texttt{Bern}(1-p)$ is pulled with probability $\pi_D(a_1) = p$. When $p$ is small, this corresponds to the better arm only being pulled occasionally. Under this setup, $\pi_{\text{RCSL}}(a_1|r=1) = \pi_{\text{RCSL}}(a_2|r=1) = 0.5$, which is highly suboptimal compared to always pulling the optimal arm $a_1$ with reward $\texttt{Bern}(1-p)$ for $p < 0.5$.
\paragraph{Results.} We train tabular \method and baselines on 1000 samples where the superior arm with $r\sim\texttt{Bern}(1-p)$ is pulled with probability $p$ for $p\in\{0.1,...,0.5\}$. Figure~\ref{fig:bandit} (right) shows that RCSL and percentage BC (filtered by $r=1$) always result in policies that are indifferent in the arms, whereas \method is able to recover the Bayes-optimal performance (dotted line) for all $p$ values considered. Future VAE performs similarly to \method for small $p$ values, but is sensitive to the KL regularization coefficient when $p$ is close to $0.5$. 

\subsection{Evaluating Stochastic Transitions in FrozenLake}
\paragraph{FrozenLake.} Next, we consider the FrozenLake environment with stochastic transitions where the agent taking an action has probability $p$ of moving in the intended direction, and probability $0.5 \cdot (1-p)$ of slipping to either of the two sides of the intended direction. We collect $100$ trajectories of length $100$ using a DQN policy trained in the original environment  ($p=\frac{1}{3}$) which achieves an average return of $0.7$, and vary $p$ during data collection and evaluation to test different levels of stochasticity. We also include uniform actions with probability $\epsilon$ to lower the performance of the offline data so that BC is highly suboptimal.

\begin{figure}[b]
    \centering
    \includegraphics[width=\linewidth]{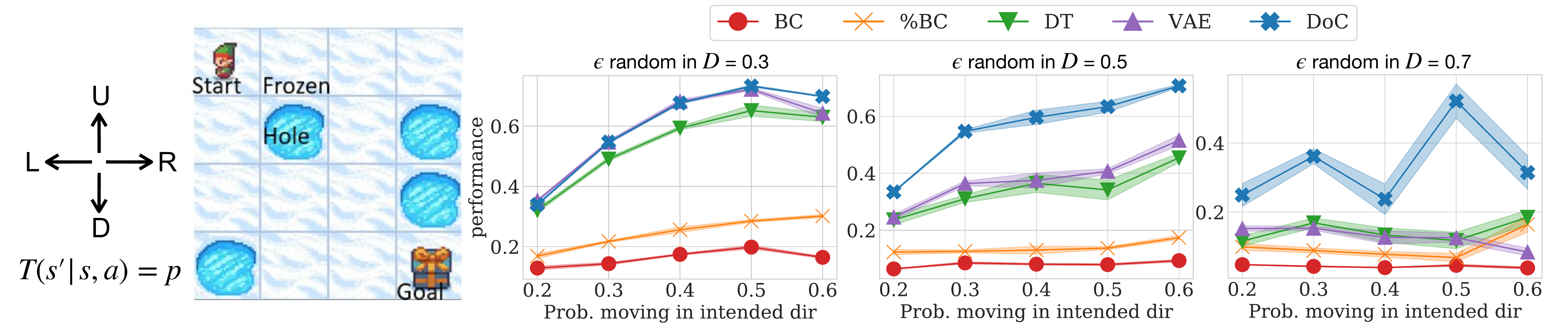}
    \caption{[Left] Visualization of the stochastic FrozenLake task. The agent has a probability $p$ of moving in the intended direction and $1-p$ of slipping to either sides. [Right] Average performance (across 5 seeds) of \method and baselines on FrozenLake with different levels of stochasticity ($p$) and offline dataset quality ($\epsilon$). \method outperforms DT and future VAE, where the gain is more salient when the offline data is less optimal ($\epsilon=0.5$ and $\epsilon=0.7$).}
    \label{fig:lake}
\end{figure}

\paragraph{Results.} Figure~\ref{fig:lake} presents the visualization (left) and results (right) for this task. When the offline data is closer to being expert ($\epsilon=0.3$), DT, future \vae, and \method perform similarly with better performance in more deterministic environments. As the offline dataset becomes more suboptimal ($\epsilon = 0.5$), \method starts to dominate across all levels of transition stochasticity. When the offline data is highly suboptimal ($\epsilon=0.7$), DT and future \vae has little advantage over BC, whereas \method continues to learn policies with reasonable performance.

\subsection{Evaluating Stochastic Gym MuJoCo}
\paragraph{Environments.} We now consider a set of Gym MuJoCo environments including Reacher, Hopper, HalfCheetah, and Humanoid. We additionally consider AntMaze from D4RL~\citep{fu2020d4rl}. These environments are deterministic by default, which we modify by introducing time-correlated Gaussian noise to the actions before inputing the action into the physics simulator during data collection and evaluation for all but AntMaze environments. Specifically, the Gaussian noise we introduce to the actions has 0 mean and standard deviation of the form $(1 - e^{-0.01 \cdot t}) \cdot \sin(t) \cdot \sigma$ where $t$ is the step number and $\sigma\in[0,1]$. For AntMaze where the dataset has already been collected in the deterministic environment by D4RL, we add gaussian noise with 0.1 standard deviation to the reward uniformly with probability 0.1 (both to the dataset and during evaluation).

\begin{figure}[t]
    \centering
    \includegraphics[width=\linewidth]{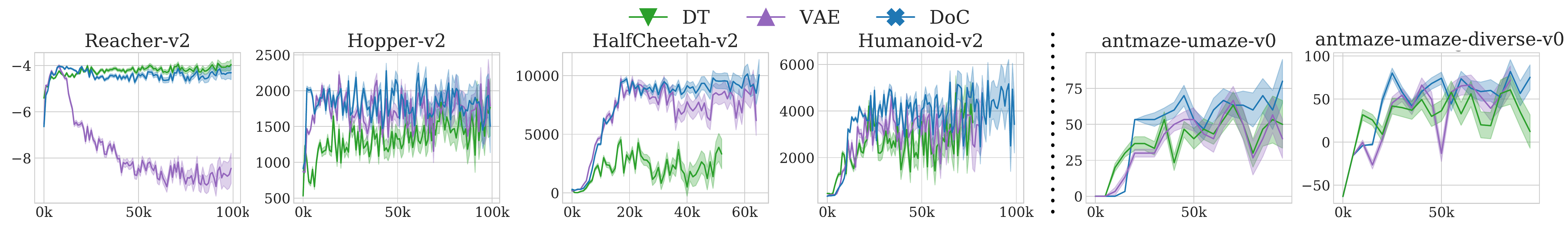}
    \caption{Average performance (across 5 seeds) of \method and baselines on modified stochastic Gym MuJoCo and AntMaze tasks. \method and future VAE generally provide benefits over DT, where \method provide more benefits on harder tasks such as Humanoid. Future \vae can be sensitive to the KL coefficient $\beta$, which can result in the failure mode shown in Reacher-v2 if not tuned properly.}
    \label{fig:mujoco}
\end{figure}

\paragraph{Results.} Figure~\ref{fig:mujoco} shows the average performance (across 5 seeds) of DT, future VAE, and \method on these stochastic environments. Both future VAE and \method generally provide benefits over DT, where the benefit of \method is more salient in harder environments such as HalfCheetah and Humanoid. We found future \vae to be sensitive to the $\beta$ hyperparameter, and simply using $\beta=1$ can result in the falure case as shown in Reacher-v2.

\section{Conclusion}
Despite the empirical promise of return- or future-conditioned supervised learning (RCSL) with large transformer architectures, environment stochasticity hampers the application of supervised learning to sequential decision making. To address this issue, we proposed to augment supervised learning with the dichotomy of control principle (\method), guiding a supervised policy to only control the controllable (actions). Theoretically, \method learns \emph{consistent} policies,  guaranteeing that they achieve the future or return they are conditioned on. Empirically, \method outperforms RCSL in highly stochastic environments. While \method still falls short in addressing other RL challenges such as `stitching' (i.e., composing sub-optimal trajectories), we hope that dichotomy of control serves as a stepping stone in solving sequential decision making with large-scale supervised learning.


\subsubsection*{Acknowledgments}
Thanks to George Tucker for reviewing draft versions of this manuscript. Thanks to Marc Bellemare for help with derivations. Thanks to David Brandfonbrener and Keiran Paster for discussions around stochastic environments. We gratefully acknowledges the support of a Canada CIFAR AI Chair, NSERC and Amii, and support from Berkeley BAIR industrial consortion.

\bibliography{iclr2023_conference}

\begin{thebibliography}{44}
\providecommand{\natexlab}[1]{#1}
\providecommand{\url}[1]{\texttt{#1}}
\expandafter\ifx\csname urlstyle\endcsname\relax
  \providecommand{\doi}[1]{doi: #1}\else
  \providecommand{\doi}{doi: \begingroup \urlstyle{rm}\Url}\fi

\bibitem[Agarwal et~al.(2020)Agarwal, Schuurmans, and
  Norouzi]{agarwal2020optimistic}
Rishabh Agarwal, Dale Schuurmans, and Mohammad Norouzi.
\newblock An optimistic perspective on offline reinforcement learning.
\newblock In \emph{International Conference on Machine Learning}, pp.\
  104--114. PMLR, 2020.

\bibitem[Ajay et~al.(2020)Ajay, Kumar, Agrawal, Levine, and
  Nachum]{ajay2020opal}
Anurag Ajay, Aviral Kumar, Pulkit Agrawal, Sergey Levine, and Ofir Nachum.
\newblock Opal: Offline primitive discovery for accelerating offline
  reinforcement learning.
\newblock \emph{arXiv preprint arXiv:2010.13611}, 2020.

\bibitem[Baker et~al.(2022)Baker, Akkaya, Zhokhov, Huizinga, Tang, Ecoffet,
  Houghton, Sampedro, and Clune]{baker2022video}
Bowen Baker, Ilge Akkaya, Peter Zhokhov, Joost Huizinga, Jie Tang, Adrien
  Ecoffet, Brandon Houghton, Raul Sampedro, and Jeff Clune.
\newblock Video pretraining (vpt): Learning to act by watching unlabeled online
  videos.
\newblock \emph{arXiv preprint arXiv:2206.11795}, 2022.

\bibitem[Brandfonbrener et~al.(2022)Brandfonbrener, Bietti, Buckman, Laroche,
  and Bruna]{brandfonbrener2022does}
David Brandfonbrener, Alberto Bietti, Jacob Buckman, Romain Laroche, and Joan
  Bruna.
\newblock When does return-conditioned supervised learning work for offline
  reinforcement learning?
\newblock \emph{arXiv preprint arXiv:2206.01079}, 2022.

\bibitem[Brockman et~al.(2016)Brockman, Cheung, Pettersson, Schneider,
  Schulman, Tang, and Zaremba]{brockman2016openai}
Greg Brockman, Vicki Cheung, Ludwig Pettersson, Jonas Schneider, John Schulman,
  Jie Tang, and Wojciech Zaremba.
\newblock Openai gym.
\newblock \emph{arXiv preprint arXiv:1606.01540}, 2016.

\bibitem[Chen et~al.(2021)Chen, Lu, Rajeswaran, Lee, Grover, Laskin, Abbeel,
  Srinivas, and Mordatch]{chen2021decision}
Lili Chen, Kevin Lu, Aravind Rajeswaran, Kimin Lee, Aditya Grover, Misha
  Laskin, Pieter Abbeel, Aravind Srinivas, and Igor Mordatch.
\newblock Decision transformer: Reinforcement learning via sequence modeling.
\newblock \emph{Advances in neural information processing systems},
  34:\penalty0 15084--15097, 2021.

\bibitem[Codevilla et~al.(2018)Codevilla, M{\"u}ller, L{\'o}pez, Koltun, and
  Dosovitskiy]{codevilla2018end}
Felipe Codevilla, Matthias M{\"u}ller, Antonio L{\'o}pez, Vladlen Koltun, and
  Alexey Dosovitskiy.
\newblock End-to-end driving via conditional imitation learning.
\newblock In \emph{2018 IEEE international conference on robotics and
  automation (ICRA)}, pp.\  4693--4700. IEEE, 2018.

\bibitem[Emmons et~al.(2021)Emmons, Eysenbach, Kostrikov, and
  Levine]{emmons2021rvs}
Scott Emmons, Benjamin Eysenbach, Ilya Kostrikov, and Sergey Levine.
\newblock Rvs: What is essential for offline rl via supervised learning?
\newblock \emph{arXiv preprint arXiv:2112.10751}, 2021.

\bibitem[Eysenbach et~al.(2022)Eysenbach, Udatha, Levine, and
  Salakhutdinov]{eysenbach2022imitating}
Benjamin Eysenbach, Soumith Udatha, Sergey Levine, and Ruslan Salakhutdinov.
\newblock Imitating past successes can be very suboptimal.
\newblock \emph{arXiv preprint arXiv:2206.03378}, 2022.

\bibitem[Fan et~al.(2022)Fan, Wang, Jiang, Mandlekar, Yang, Zhu, Tang, Huang,
  Zhu, and Anandkumar]{fan2022minedojo}
Linxi Fan, Guanzhi Wang, Yunfan Jiang, Ajay Mandlekar, Yuncong Yang, Haoyi Zhu,
  Andrew Tang, De-An Huang, Yuke Zhu, and Anima Anandkumar.
\newblock Minedojo: Building open-ended embodied agents with internet-scale
  knowledge.
\newblock \emph{arXiv preprint arXiv:2206.08853}, 2022.

\bibitem[Fox et~al.(2017)Fox, Krishnan, Stoica, and Goldberg]{fox2017multi}
Roy Fox, Sanjay Krishnan, Ion Stoica, and Ken Goldberg.
\newblock Multi-level discovery of deep options.
\newblock \emph{arXiv preprint arXiv:1703.08294}, 2017.

\bibitem[Fu et~al.(2020)Fu, Kumar, Nachum, Tucker, and Levine]{fu2020d4rl}
Justin Fu, Aviral Kumar, Ofir Nachum, George Tucker, and Sergey Levine.
\newblock D4rl: Datasets for deep data-driven reinforcement learning.
\newblock \emph{arXiv preprint arXiv:2004.07219}, 2020.

\bibitem[Fujimoto et~al.(2019)Fujimoto, Meger, and Precup]{fujimoto2019off}
Scott Fujimoto, David Meger, and Doina Precup.
\newblock Off-policy deep reinforcement learning without exploration.
\newblock In \emph{International conference on machine learning}, pp.\
  2052--2062. PMLR, 2019.

\bibitem[Furuta et~al.(2021)Furuta, Matsuo, and Gu]{furuta2021generalized}
Hiroki Furuta, Yutaka Matsuo, and Shixiang~Shane Gu.
\newblock Generalized decision transformer for offline hindsight information
  matching.
\newblock \emph{arXiv preprint arXiv:2111.10364}, 2021.

\bibitem[Ghosh et~al.(2019)Ghosh, Gupta, Reddy, Fu, Devin, Eysenbach, and
  Levine]{ghosh2019learning}
Dibya Ghosh, Abhishek Gupta, Ashwin Reddy, Justin Fu, Coline Devin, Benjamin
  Eysenbach, and Sergey Levine.
\newblock Learning to reach goals via iterated supervised learning.
\newblock \emph{arXiv preprint arXiv:1912.06088}, 2019.

\bibitem[Goyal et~al.(2019)Goyal, Lamb, Hoffmann, Sodhani, Levine, Bengio, and
  Sch{\"o}lkopf]{goyal2019recurrent}
Anirudh Goyal, Alex Lamb, Jordan Hoffmann, Shagun Sodhani, Sergey Levine,
  Yoshua Bengio, and Bernhard Sch{\"o}lkopf.
\newblock Recurrent independent mechanisms.
\newblock \emph{arXiv preprint arXiv:1909.10893}, 2019.

\bibitem[Haarnoja et~al.(2018)Haarnoja, Zhou, Abbeel, and
  Levine]{haarnoja2018soft}
Tuomas Haarnoja, Aurick Zhou, Pieter Abbeel, and Sergey Levine.
\newblock Soft actor-critic: Off-policy maximum entropy deep reinforcement
  learning with a stochastic actor.
\newblock In \emph{International conference on machine learning}, pp.\
  1861--1870. PMLR, 2018.

\bibitem[Hakhamaneshi et~al.(2021)Hakhamaneshi, Zhao, Zhan, Abbeel, and
  Laskin]{hakhamaneshi2021hierarchical}
Kourosh Hakhamaneshi, Ruihan Zhao, Albert Zhan, Pieter Abbeel, and Michael
  Laskin.
\newblock Hierarchical few-shot imitation with skill transition models.
\newblock \emph{arXiv preprint arXiv:2107.08981}, 2021.

\bibitem[Kingma \& Welling(2013)Kingma and Welling]{kingma2013auto}
Diederik~P Kingma and Max Welling.
\newblock Auto-encoding variational bayes.
\newblock \emph{arXiv preprint arXiv:1312.6114}, 2013.

\bibitem[Kipf et~al.(2019)Kipf, Li, Dai, Zambaldi, Sanchez-Gonzalez,
  Grefenstette, Kohli, and Battaglia]{kipf2019compile}
Thomas Kipf, Yujia Li, Hanjun Dai, Vinicius Zambaldi, Alvaro Sanchez-Gonzalez,
  Edward Grefenstette, Pushmeet Kohli, and Peter Battaglia.
\newblock Compile: Compositional imitation learning and execution.
\newblock In \emph{International Conference on Machine Learning}, pp.\
  3418--3428. PMLR, 2019.

\bibitem[Krishnan et~al.(2017)Krishnan, Fox, Stoica, and
  Goldberg]{krishnan2017ddco}
Sanjay Krishnan, Roy Fox, Ion Stoica, and Ken Goldberg.
\newblock Ddco: Discovery of deep continuous options for robot learning from
  demonstrations.
\newblock In \emph{Conference on robot learning}, pp.\  418--437. PMLR, 2017.

\bibitem[Kumar et~al.(2019)Kumar, Peng, and Levine]{kumar2019reward}
Aviral Kumar, Xue~Bin Peng, and Sergey Levine.
\newblock Reward-conditioned policies.
\newblock \emph{arXiv preprint arXiv:1912.13465}, 2019.

\bibitem[Kumar et~al.(2020)Kumar, Zhou, Tucker, and
  Levine]{kumar2020conservative}
Aviral Kumar, Aurick Zhou, George Tucker, and Sergey Levine.
\newblock Conservative q-learning for offline reinforcement learning.
\newblock \emph{Advances in Neural Information Processing Systems},
  33:\penalty0 1179--1191, 2020.

\bibitem[Kumar et~al.(2021)Kumar, Hong, Singh, and Levine]{kumar2021should}
Aviral Kumar, Joey Hong, Anikait Singh, and Sergey Levine.
\newblock Should i run offline reinforcement learning or behavioral cloning?
\newblock In \emph{International Conference on Learning Representations}, 2021.

\bibitem[Lee et~al.(2022)Lee, Nachum, Yang, Lee, Freeman, Xu, Guadarrama,
  Fischer, Jang, Michalewski, et~al.]{lee2022multi}
Kuang-Huei Lee, Ofir Nachum, Mengjiao Yang, Lisa Lee, Daniel Freeman, Winnie
  Xu, Sergio Guadarrama, Ian Fischer, Eric Jang, Henryk Michalewski, et~al.
\newblock Multi-game decision transformers.
\newblock \emph{arXiv preprint arXiv:2205.15241}, 2022.

\bibitem[Lynch et~al.(2020)Lynch, Khansari, Xiao, Kumar, Tompson, Levine, and
  Sermanet]{lynch2020learning}
Corey Lynch, Mohi Khansari, Ted Xiao, Vikash Kumar, Jonathan Tompson, Sergey
  Levine, and Pierre Sermanet.
\newblock Learning latent plans from play.
\newblock In \emph{Conference on Robot Learning}, pp.\  1113--1132. PMLR, 2020.

\bibitem[Mnih et~al.(2013)Mnih, Kavukcuoglu, Silver, Graves, Antonoglou,
  Wierstra, and Riedmiller]{mnih2013playing}
Volodymyr Mnih, Koray Kavukcuoglu, David Silver, Alex Graves, Ioannis
  Antonoglou, Daan Wierstra, and Martin Riedmiller.
\newblock Playing atari with deep reinforcement learning.
\newblock \emph{arXiv preprint arXiv:1312.5602}, 2013.

\bibitem[Paster et~al.(2020)Paster, McIlraith, and Ba]{paster2020planning}
Keiran Paster, Sheila~A McIlraith, and Jimmy Ba.
\newblock Planning from pixels using inverse dynamics models.
\newblock \emph{arXiv preprint arXiv:2012.02419}, 2020.

\bibitem[Paster et~al.(2022)Paster, McIlraith, and Ba]{paster2022you}
Keiran Paster, Sheila McIlraith, and Jimmy Ba.
\newblock You can't count on luck: Why decision transformers fail in stochastic
  environments.
\newblock \emph{arXiv preprint arXiv:2205.15967}, 2022.

\bibitem[Pertsch et~al.(2020)Pertsch, Lee, and Lim]{pertsch2020accelerating}
Karl Pertsch, Youngwoon Lee, and Joseph~J Lim.
\newblock Accelerating reinforcement learning with learned skill priors.
\newblock \emph{arXiv preprint arXiv:2010.11944}, 2020.

\bibitem[Puterman(2014)]{puterman2014markov}
Martin~L Puterman.
\newblock \emph{Markov decision processes: discrete stochastic dynamic
  programming}.
\newblock John Wiley \& Sons, 2014.

\bibitem[Reed et~al.(2022)Reed, Zolna, Parisotto, Colmenarejo, Novikov,
  Barth-Maron, Gimenez, Sulsky, Kay, Springenberg, et~al.]{reed2022generalist}
Scott Reed, Konrad Zolna, Emilio Parisotto, Sergio~Gomez Colmenarejo, Alexander
  Novikov, Gabriel Barth-Maron, Mai Gimenez, Yury Sulsky, Jackie Kay,
  Jost~Tobias Springenberg, et~al.
\newblock A generalist agent.
\newblock \emph{arXiv preprint arXiv:2205.06175}, 2022.

\bibitem[Reid et~al.(2022)Reid, Yamada, and Gu]{reid2022can}
Machel Reid, Yutaro Yamada, and Shixiang~Shane Gu.
\newblock Can wikipedia help offline reinforcement learning?
\newblock \emph{arXiv preprint arXiv:2201.12122}, 2022.

\bibitem[Schmidhuber(2019)]{schmidhuber2019reinforcement}
Juergen Schmidhuber.
\newblock Reinforcement learning upside down: Don't predict rewards--just map
  them to actions.
\newblock \emph{arXiv preprint arXiv:1912.02875}, 2019.

\bibitem[Shankar \& Gupta(2020)Shankar and Gupta]{shankar2020learning}
Tanmay Shankar and Abhinav Gupta.
\newblock Learning robot skills with temporal variational inference.
\newblock In \emph{International Conference on Machine Learning}, pp.\
  8624--8633. PMLR, 2020.

\bibitem[Shapiro(2014)]{shapiro2014wrote}
Fred~R Shapiro.
\newblock Who wrote the serenity prayer?
\newblock \emph{The Chronicle Review}, 28, 2014.

\bibitem[Singh et~al.(2020)Singh, Liu, Zhou, Yu, Rhinehart, and
  Levine]{singh2020parrot}
Avi Singh, Huihan Liu, Gaoyue Zhou, Albert Yu, Nicholas Rhinehart, and Sergey
  Levine.
\newblock Parrot: Data-driven behavioral priors for reinforcement learning.
\newblock \emph{arXiv preprint arXiv:2011.10024}, 2020.

\bibitem[Srivastava et~al.(2019)Srivastava, Shyam, Mutz, Ja{\'s}kowski, and
  Schmidhuber]{srivastava2019training}
Rupesh~Kumar Srivastava, Pranav Shyam, Filipe Mutz, Wojciech Ja{\'s}kowski, and
  J{\"u}rgen Schmidhuber.
\newblock Training agents using upside-down reinforcement learning.
\newblock \emph{arXiv preprint arXiv:1912.02877}, 2019.

\bibitem[{\v{S}}trupl et~al.(2022){\v{S}}trupl, Faccio, Ashley, Schmidhuber,
  and Srivastava]{vstrupl2022upside}
Miroslav {\v{S}}trupl, Francesco Faccio, Dylan~R Ashley, J{\"u}rgen
  Schmidhuber, and Rupesh~Kumar Srivastava.
\newblock Upside-down reinforcement learning can diverge in stochastic
  environments with episodic resets.
\newblock \emph{arXiv preprint arXiv:2205.06595}, 2022.

\bibitem[Venuto et~al.(2021)Venuto, Lau, Precup, and Nachum]{venuto2021policy}
David Venuto, Elaine Lau, Doina Precup, and Ofir Nachum.
\newblock Policy gradients incorporating the future.
\newblock \emph{arXiv preprint arXiv:2108.02096}, 2021.

\bibitem[Villaflor et~al.(2022)Villaflor, Huang, Pande, Dolan, and
  Schneider]{villaflor2022addressing}
Adam~R Villaflor, Zhe Huang, Swapnil Pande, John~M Dolan, and Jeff Schneider.
\newblock Addressing optimism bias in sequence modeling for reinforcement
  learning.
\newblock In \emph{International Conference on Machine Learning}, pp.\
  22270--22283. PMLR, 2022.

\bibitem[Wang et~al.(2021)Wang, Lee, Hakhamaneshi, Abbeel, and
  Laskin]{wang2021skill}
Xiaofei Wang, Kimin Lee, Kourosh Hakhamaneshi, Pieter Abbeel, and Michael
  Laskin.
\newblock Skill preferences: Learning to extract and execute robotic skills
  from human feedback.
\newblock \emph{arXiv preprint arXiv:2108.05382}, 2021.

\bibitem[Wu et~al.(2019)Wu, Tucker, and Nachum]{wu2019behavior}
Yifan Wu, George Tucker, and Ofir Nachum.
\newblock Behavior regularized offline reinforcement learning.
\newblock \emph{arXiv preprint arXiv:1911.11361}, 2019.

\bibitem[Zhou et~al.(2020)Zhou, Bajracharya, and Held]{zhou2020plas}
Wenxuan Zhou, Sujay Bajracharya, and David Held.
\newblock Plas: Latent action space for offline reinforcement learning.
\newblock \emph{arXiv preprint arXiv:2011.07213}, 2020.

\end{thebibliography}
\bibliographystyle{iclr2023_conference}

\appendix
\clearpage
\begin{center}
{\huge Appendix}
\end{center}

\section{Proof of Theorem~\ref{thm:consistency}}
\label{sec:proof}
The proof relies on the following lemma, showing that the MI constraints ensure that the observed rewards and dynamics conditioned on $z$ in the training data are equal to the rewards and dynamics of the environment. 
\begin{lemma}
\label{lem:mi}
Suppose $\method$ yields $q$ satisfying the MI constraints:
\begin{equation}
    \mathrm{MI}(r_t;z|\tau_{0:t-1},s_t,a_t)=\mathrm{MI}(s_{t+1};z|\tau_{0:t-1},s_t,a_t)=0,
\end{equation}
for all $\tau_{0:t-1},s_t,a_t$ with $\prob[\tau_{0:t-1},s_t,a_t|\Dset] > 0$.
Then under Assumption~\ref{ass:data-env},
\begin{align}
    \prob\left[\hat{r}=r_t~|~\tau_{0:t-1},s_t,a_t,z,\Dset\right] &= \Reward(\hat{r}_t ~|~\tau_{0:t-1}, s_t, a_t), \\
    \prob\left[\hat{s}_{t+1}=s_{t+1}~|~\tau_{0:t-1},s_t,a_t,z,\Dset\right] &= \Trans(\hat{s}_{t+1} ~|~ \tau_{0:t-1}, s_t, a_t),
\end{align}
for all $\tau_{0:t-1},s_t,a_t,z$ and $\hat{r},\hat{s}_{t+1}$, as long as $\prob[\tau_{0:t-1},s_t,a_t,z|\Dset]>0$. 
%Note that we use the notation $\prob\left[E|~\tau_{0:t-1},s_t,a_t,z,\Dset\right]$ to denote the probability density of event $E$ when $\tau$ sampled from the dataset $\Dset$ and $z$ sampled from $q(\tau)$, conditioned on the first $t$ steps of $\tau$ being equal to $\tau_{0:t-1},s_t,a_t$.
\end{lemma}
\proof{
We show the derivations relevant to reward, with those for next-state being analogous. We start with the definition of mutual information:
\begin{equation}
    \mathrm{MI}(r_t;z|\tau_{0:t-1},s_t,a_t) = \E_{(r_t,z)\sim \prob[\cdot|\tau_{0:t-1},s_t,a_t,\Dset]}\left[\log \frac{\prob\left[r_t|\tau_{0:t-1},s_t,a_t,z,\Dset\right]}{\prob\left[r_t|\tau_{0:t-1},s_t,a_t, \Dset\right]}\right]
\end{equation}
\begin{equation}
    = \E_{z\sim \prob[\cdot|\tau_{0:t-1},s_t,a_t,\Dset]}\left[\dkl(\prob[r|\tau_{0:t-1},s_t,a_t,z,\Dset]\|\prob[r|\tau_{0:t-1},s_t,a_t, \Dset])\right].
\end{equation}
The KL divergence is a nonnegative quantity, and it is zero only when the two input distributions are equal. Thus, the constraint $\mathrm{MI}(r_t;z|\tau_{0:t-1},s_t,a_t)=0$ implies,
\begin{equation}
    \prob\left[r|\tau_{0:t-1},s_t,a_t,z,\Dset\right] = \prob[r|\tau_{0:t-1},s_t,a_t, \Dset],
\end{equation}
for all $\tau_{0:t-1},s_t,a_t,z$ with $\prob[z|\tau_{0:t-1},s_t,a_t,\Dset]>0$.
From Assumption~\ref{ass:data-env} we know
\begin{equation}
    \prob[r|\tau_{0:t-1},s_t,a_t, \Dset] = \Reward(r|\tau_{0:t-1},s_t,a_t),
\end{equation}
and so we immediately have the desired result.
}

We will further employ the following lemma, which takes us most of the way to proving Theorem~\ref{thm:consistency}:
\begin{lemma}
\label{lem:visitation}
Suppose \method yields $\pi, q$ with $q$ satisfying the MI constraints:
\begin{equation}
    \mathrm{MI}(r_t;z|\tau_{0:t-1},s_t,a_t)=\mathrm{MI}(s_{t+1};z|\tau_{0:t-1},s_t,a_t)=0,
\end{equation}
for all $\tau_{0:t-1},s_t,a_t$ with $\prob[\tau_{0:t-1},s_t,a_t|\Dset] > 0$.
Then under Assumptions~\ref{ass:data-env} and~\ref{ass:bayes}, we have
\begin{equation}
    \prob\left[\tau~|~z,\Dset\right] = \prob\left[\tau~|~\pi_z,\mdp\right],
\end{equation}
for all $\tau$ and all $z$ with $\prob[z|q,\Dset]>0$. 
\end{lemma}
\proof{
We may write the probability $\prob\left[\tau~|~z,\Dset\right]$ as,
\begin{multline}
    \prob\left[\tau ~|~ z, \Dset\right] = \prod_{t=0}^H \prob\left[a_t ~|~ \tau_{0:t-1}, s_t, z, \Dset\right]  \\
    \cdot \prod_{t=0}^H \prob\left[r_t ~|~ \tau_{0:t-1}, s_t, a_t, z, \Dset\right]  \\
    \cdot \prod_{t=0}^{H-1} \prob\left[s_{t+1} ~|~ \tau_{0:t-1}, s_t, a_t, z, \Dset\right].
\end{multline}
\paragraph{Case 1:} We begin by considering the case of $\tau$ satisfying $\prob\left[\tau ~|~ z, \Dset\right]>0$. 
For such a $\tau$, by Assumption~\ref{ass:bayes} we may write the first probability above as
\begin{equation}
    \prob\left[a_t ~|~ \tau_{0:t-1}, s_t, z, \Dset\right] = \pi_z(a_t | \tau_{0:t-1}, s_t).
\end{equation}
Moreover, by Lemma~\ref{lem:mi} we may write the second and third probabilities as
\begin{align}
    \prob\left[r_t ~|~ \tau_{0:t-1}, s_t, a_t, z, \Dset\right] &= \Reward(r_t | \tau_{0:t-1}, s_t, a_t) \\
    \prob\left[s_{t+1} ~|~ \tau_{0:t-1}, s_t, a_t, z, \Dset\right] &= \Trans(s_{t+1} | \tau_{0:t-1}, s_t, a_t).
\end{align}
Therefore, for any $\tau$ with $\prob\left[\tau ~|~ z, \Dset\right]>0$ we have,
\begin{align}
    \prob\left[\tau ~|~ z, \Dset\right] &= \prod_{t=0}^H \pi_z(a_t | \tau_{0:t-1}, s_t)  
    \cdot \prod_{t=0}^H \Reward(r_t | \tau_{0:t-1}, s_t, a_t)
    \cdot \prod_{t=0}^{H-1} \Trans(s_{t+1} | \tau_{0:t-1}, s_t, a_t) \nonumber \\
    &=  \prob\left[\tau ~|~ \pi_z, \mdp\right].
\end{align}

\paragraph{Case 2:} To handle the case of $\prob\left[\tau~|~z,\Dset\right]=0$ we will show that $\prob\left[\tau_{0:t}~|~z,\Dset\right]=0$ implies $\prob\left[\tau_{0:t}~|~\pi_z,\mdp\right]=0$ by induction on $t$.
The base case of $t=-1$ is trivial. For $t>-1$, we may write,
\begin{multline}
    \label{eq:induction}
    \prob\left[\tau_{0:t}~|~z,\Dset\right] =
    \prob\left[\tau_{0:t-1}~|~z,\Dset\right]\cdot  \prob\left[s_t~|~\tau_{0:t-2},s_{t-1},a_{t-1},z,\Dset\right]\cdot \prob\left[a_t~|~\tau_{0:t-1},s_t,z,\Dset\right]\cdot \\ \prob\left[r_t~|~\tau_{0:t-1},s_t,a_t,z,\Dset\right],
\end{multline}
\begin{multline}
    \prob\left[\tau_{0:t}~|~\pi_z,\mdp\right] =
    \prob\left[\tau_{0:t-1}~|~\pi_z,\mdp\right]\cdot  \Trans(s_t|\tau_{0:t-2},s_{t-1},a_{t-1})\cdot \pi_z(a_t|\tau_{0:t-1},s_t)\cdot \\ \Reward(r_t|\tau_{0:t-1},s_t,a_t).
\end{multline}
Suppose, for the sake of contradiction, that $\prob\left[\tau_{0:t}~|~z,\Dset\right]=0$ while $\prob\left[\tau_{0:t}~|~\pi_z,\mdp\right]>0$. 
By the inductive hypothesis, $\prob\left[\tau_{0:t-1}~|~z,\Dset\right]>0$.
Thus, by Lemma~\ref{lem:mi} we must have
\begin{equation}
    \prob\left[s_t~|~\tau_{0:t-2},s_{t-1},a_{t-1},z,\Dset\right] = \Trans(s_t|\tau_{0:t-2},s_{t-1},a_{t-1}),
\end{equation}
and so $\Trans(s_t|\tau_{0:t-2},s_{t-1},a_{t-1})>0$ implies that $\prob\left[s_t~|~\tau_{0:t-2},s_{t-1},a_{t-1},z,\Dset\right]>0$.
Thus, by Assumption~\ref{ass:bayes} we must have
\begin{equation}
    \prob\left[a_t~|~\tau_{0:t-1},s_t,z,\Dset\right] = \pi_z(a_t|\tau_{0:t-1},s_t),
\end{equation}
and so $\pi_z(a_t|\tau_{0:t-1},s_t)>0$ implies that $\prob\left[a_t~|~\tau_{0:t-1},s_t,z,\Dset\right]>0$.
Lastly, by Lemma~\ref{lem:mi} we must have
\begin{equation}
    \prob\left[r_t~|~\tau_{0:t-1},s_t,a_t,z,\Dset\right] = \Reward(r_t|\tau_{0:t-1},s_t,a_t),
\end{equation}
and so $\Reward(r_t|\tau_{0:t-1},s_t,a_t)>0$ implies that $\prob\left[r_t~|~\tau_{0:t-1},s_t,a_t,z,\Dset\right]>0$. 
Altogether, we find that each of the three terms on the RHS of~\Eqref{eq:induction} is strictly positive and so $\prob\left[\tau_{0:t}~|~z,\Dset\right]>0$; contradiction.
}

\subsection{Theorem Proof}
\label{sec:thm-proof}
We are now prepared to prove Theorem~\ref{thm:consistency}.

Using Assumption~\ref{ass:bayes}, we can express $V(z)$ as,
\begin{align}
    V(z) &= \int \prob\left[\hat{\tau} = \tau ~|~ z, \Dset\right] \cdot \return(\hat{\tau})~d\hat{\tau}.
\end{align}
%where we use $1[E]$ to denote $1$ if $E$ is true and $0$ otherwise.
By Lemma~\ref{lem:visitation} we have,
\begin{align}
     V(z) &= \int \prob\left[\hat{\tau}=
     \tau ~|~ z, \Dset\right] \cdot \return(\hat{\tau})~d\hat{\tau} \\
     &= \int \prob\left[\hat{\tau} = \tau ~|~ \pi_z, \mdp \right] \cdot \return(\hat{\tau})~d\hat{\tau} \\
     &= V_\mdp(\pi_z),
\end{align}
as desired.

%\subsection{Relaxing the Conditions of Theorem~\ref{thm:consistency}}
%As clear from the derivations in Section~\ref{sec:thm-proof} above, our proof hinges on showing that $\prob[\tau~|~z,\Dset]=\prob[\tau~|~\pi_z,\mdp]$ for all $z,\tau$. While the exact equivalence of these two distributions requires the strict conditions of Theorem~\ref{thm:consistency}, one may alternatively aim to show
%\begin{equation}
%    \dtv(\prob[\cdot|~z,\Dset] \| \prob[\cdot|~\pi_z,\mdp]) \le \epsilon,
%\end{equation}
%which would induce an $\epsilon$-approximate form of consistency between $V,\pi$.

%The advantage of relaxing the strict requirements on consistency allows one to relax the conditions of Theorem~\ref{thm:consistency}, namely Assumption~\ref{ass:bayes} and the exact MI constraints on $q$. For relaxed conditions, one may derive a result analogous to Theorem~\ref{thm:consistency} by following the same logic and using standard techniques (see~\citet{nachum2021provable,schulman2015trust}) for bounding $\dtv(\prob[\cdot|~z,\Dset] \| \prob[\cdot|~\pi_z,\mdp])$ in terms of bounds on
%\begin{align}
%    \dtv(\prob\left[\hat{a}_t = a_t ~|~ \hat{\tau}_{0:t-1} = \tau_{0:t-1}, s_t, z, \Dset\right] &~\|~ \pi_z(\hat{a}_t | \hat{\tau}_{0:t-1}, s_t)), \\
%    \dtv(\prob\left[\hat{r}_t = r_t ~|~ \hat{\tau}_{0:t-1} = \tau_{0:t-1}, s_t, a_t, z, \Dset\right] &~\|~ \Reward(\hat{r}_t | \hat{\tau}_{0:t-1}, s_t, a_t), \\
%    \dtv(\prob\left[\hat{s}_{t+1} = s_{t+1} ~|~ \hat{\tau}_{0:t-1} = \tau_{0:t-1}, s_t, a_t, z, \Dset\right] &~\|~ \Trans(\hat{s}_{t+1} | \hat{\tau}_{0:t-1}, s_t, a_t)).
%\end{align}

\section{Proof of Theorem~\ref{thm:consistency2}} % in Markovian Environments}
\label{sec:markov-policies}
We begin by proving a result under stricter conditions, namely, when the MI constraints retain the conditioning on history. % The proof of this result will illuminate the key idea for proving the more general result in Theorem~\ref{thm:consistency2}.
\begin{lemma}
\label{lem:consistency2a}
Suppose \method yields $\pi, V, q$ with $q$ satisfying the MI constraints:
\begin{equation}
    \mathrm{MI}(r_t;z|\tau_{0:t-1},s_t,a_t)=\mathrm{MI}(s_{t+1};z|\tau_{0:t-1},s_t,a_t)=0,
\end{equation}
for all $\tau_{0:t-1},s_t,a_t$ with $\prob[\tau_{0:t-1},s_t,a_t|\Dset] > 0$.
Then under Assumptions~\ref{ass:data-env},~\ref{ass:markov}, and~\ref{ass:bayes2}, $V$ and $\pi$ are consistent for any $z$ with $\prob[z|q,\Dset] >0$.
\end{lemma}
\proof{
Let
\begin{equation}
    \pi_z^{\mathrm{hist}}(\hat{a}~|~\tau_{0:t-1},s_t) = \prob\left[\hat{a} = a_t ~|~ \tau_{0:t-1}, s_t, z, \Dset\right].
\end{equation}
By Lemma~\ref{lem:visitation} and Theorem~\ref{thm:consistency} we have
\begin{equation}
    \prob\left[\tau ~|~ z, \Dset\right] =  \prob\left[\tau ~|~ \pi_z^\mathrm{hist}, \mdp\right],
    \label{eq:history-occupancies}
\end{equation}
for all $\tau$ and
\begin{equation}
    V(z) = V_\mdp(\pi_z^\mathrm{hist}),
\end{equation}
for all $z$ with $\prob[z~|~q,\Dset]>0$.

It is left to show that $V_\mdp(\pi_z^\mathrm{hist}) = V_\mdp(\pi_z)$.
To do so, we invoke Theorem 5.5.1 in~\citet{puterman2014markov}, which states that, for any history-dependent policy, there exists a Markov policy such that the state-action visitation occupancies of the two policies are equal (and, accordingly, their values are equal). In other words, there exists a Markov policy $\tilde{\pi}_z$ such that
\begin{equation}
    \label{eq:puterman}
    \prob\left[\hat{s}=s_t,\hat{a}=a_t~|~\pi_z^\mathrm{hist}, \mdp\right] = \prob\left[\hat{s}=s_t,\hat{a}=a_t~|~\tilde{\pi}_z, \mdp\right],
\end{equation}
for all $t,\hat{s},\hat{a}$, and
\begin{equation}
    V_\mdp(\pi_z^\mathrm{hist}) = V_\mdp(\tilde{\pi}_z).
\end{equation}
To complete the proof, we show that $\tilde{\pi}_z = \pi_z$. By~\Eqref{eq:history-occupancies} we have
\begin{equation}
    \label{eq:mdp-data-occupancies}
    \prob\left[\hat{s}=s_t,\hat{a}=a_t~|~\pi_z^\mathrm{hist}, \mdp\right] = \prob\left[\hat{s}=s_t,\hat{a}=a_t~|~z, \Dset\right].
\end{equation}
Thus, for any $t,\hat{s},\hat{a}$ we have
\begin{align}
    \tilde{\pi}_z(\hat{a}=a_t|\hat{s}=s_t) &= \frac{\prob\left[\hat{s}=s_t,\hat{a}=a_t~|~\tilde{\pi}_z, \mdp\right]}{\prob\left[\hat{s}=s_t~|\tilde{\pi}_z, \mdp\right]} \\
    &= \frac{\prob\left[\hat{s}=s_t,\hat{a}=a_t~|~\pi^\mathrm{hist}_z, \mdp\right]}{\prob\left[\hat{s}=s_t~|~\pi^\mathrm{hist}_z, \mdp\right]} \\
    &= \frac{\prob\left[\hat{s}=s_t,\hat{a}=a_t~|~z,\Dset\right]}{\prob\left[\hat{s}=s_t~|~z,\Dset\right]} \\
    &= \pi_z(\hat{a} = a_t | \hat{s} = s_t),
\end{align}
where the first equality is Bayes' rule, the second equality is due to~\Eqref{eq:puterman}, the third equality is due to~\Eqref{eq:mdp-data-occupancies}, and last equality is by definition of $\pi_z$ (Assumption~\ref{ass:bayes2}).
}

Before continuing to the main proof, we present the following analogue to Lemma~\ref{lem:mi}:
\begin{lemma}
\label{lem:mi2}
Suppose $\method$ yields $q$ satisfying the MI constraints:
\begin{equation}
    \mathrm{MI}(r_t;z|s_t,a_t)=\mathrm{MI}(s_{t+1};z|s_t,a_t)=0,
\end{equation}
for all $s_t,a_t$ with $\prob[s_t,a_t|\Dset] > 0$.
Then under Assumptions~\ref{ass:data-env} and~\ref{ass:markov},
\begin{align}
    \prob\left[\hat{r}=r_t~|~s_t,a_t,z,\Dset\right] &= \Reward(\hat{r}_t ~|~s_t, a_t), \\
    \prob\left[\hat{s}_{t+1}=s_{t+1}~|~s_t,a_t,z,\Dset\right] &= \Trans(\hat{s}_{t+1} ~|~ s_t, a_t),
\end{align}
for all $s_t,a_t,z$ and $\hat{r},\hat{s}_{t+1}$, as long as $\prob[s_t,a_t,z|\Dset]>0$. 
\end{lemma}
\proof{The proof is analogous to the proof of Lemma~\ref{lem:mi}.}

\subsection{Theorem Proof}
We can now tackle the proof of Theorem~\ref{thm:consistency2}.
To do so, we start by interpreting the episodes $\tau$ in the training data $\Dset$ as coming from a modified Markovian environment $\mdp^\dagger$. Specifically, we define $\mdp^\dagger$ as an environment with the same state space as $\mdp$ but with an action space consisting of tuples $(a,r,s')$, where $a$ is an action from the action space of $\mdp$, $r$ is a scalar, and $s'$ is a state from the state space of $\mdp$. We define the reward and transition functions of $\mdp^\dagger$ to be deterministic, so that the reward and next state associated with $(a,r,s')$ is $r$ and $s'$, respectively. This way, we may interpret any episode $\tau=(s_t,a_t,r_t)_{t=0}^H$ in $\mdp$ as an episode
\begin{equation}
    \tau^\dagger = (s_t, (a_t,r_t,s_{t+1}), r_t)_{t=0}^H
\end{equation}
in the modified environment $\mdp^\dagger$. Denoting $\Dset^\dagger$ as the training data distribution when interpreted in this way, we note that the MI constraints of Lemma~\ref{lem:consistency2a} hold, since rewards and transitions are deterministic. Thus, the policy $\pi^\dagger$ defined as
\begin{equation}
    \pi^\dagger((\hat{a},\hat{r},\hat{s}') | s_t,z) = \prob[(\hat{a},\hat{r},\hat{s}') = (a_t,r_t,s_{t+1}) | s_t,z,\Dset^\dagger]
\end{equation}
satisfies
\begin{equation}
    V(z) = V_{\mdp^\dagger}(\pi^\dagger_z).
\end{equation}
It is left to show that $V_{\mdp^\dagger}(\pi^\dagger_z)=V_{\mdp}(\pi_z)$. To do so, consider an episode $\tau^\dagger\sim\prob[\cdot|\pi_z^\dagger,\mdp^\dagger]$. For any single-step transition in this episode,
\begin{equation}
    (s_t, (a_t,r_t,s_{t+1}), r_t, s_{t+1}),
\end{equation}
we have, by definition of $\pi_z^\dagger$,
\begin{equation}
    \prob[\hat{a} = a_t | s_t, \pi^\dagger_z] = \prob[\hat{a}=a_t|s_t,z,\Dset^\dagger] = \pi_z(\hat{a} | s_t).
\end{equation}
In a similar vein, by definition of $\pi_z^\dagger$ and Lemma~\ref{lem:mi2} we have,
\begin{align}
    \prob[\hat{r} = r_t | s_t, a_t, \pi^\dagger_z] &= \prob[\hat{r}=r_t|s_t,a_t,z,\Dset^\dagger] = \Reward(\hat{r}|s_t,a_t), \\
    \prob[\hat{s}_{t+1} = s_{t+1} | s_t, a_t, \pi^\dagger_z] &= \prob[\hat{s}_{t+1}=s_{t+1}|s_t,a_t,z,\Dset^\dagger] = \Trans(\hat{s}_{t+1}|s_t,a_t).
\end{align}
Thus, any $\tau^\dagger=(s_t, (a_t,r_t,s_{t+1}), r_t)_{t=0}^H$ sampled from $\pi^\dagger_z, \mdp^\dagger$ can be mapped back to a $\tau=(s_t,a_t,r_t)_{t=0}^H$ in the original environment $\mdp$, where $\prob[\tau^\dagger|\pi^\dagger_z,\mdp^\dagger] = \prob[\tau|\pi_z,\mdp]$. It is clear that $\return(\tau^\dagger)=\return(\tau)$, and so we immediately have
\begin{equation}
    V_{\mdp^\dagger}(\pi^\dagger_z)=V_{\mdp}(\pi_z),
\end{equation}
as desired.

\section{Invalidity of Alternative Consistency Frameworks}
\label{sec:bad-consistency}
\citet{paster2022you} propose a similar but distinct notion of consistency compared to ours (i.e., Definition~\ref{def:consistency}), and claim that it can be achieved with stationary policies in Markovian environments. In this section, we show that this is, in fact, false, supporting the benefits of our framework.
%and calling into question %the correctness of Theorem 2.1 in~
%some of the claims in~%
%\citet{paster2022you}. 
We begin by rephrasing Theorem 2.1 of~\citet{paster2022you} using our own notation:

\paragraph{(Incorrect) Theorem 2.1 of~\citet{paster2022you}.}
\emph{
Suppose $\mdp$ is Markovian and $\Dset,q$ are given such that 
\begin{equation}
    \prob[\hat{s}_{t+1} = s_t~|~ s_t,a_t,z,\Dset] = \prob[\hat{s}_{t+1} = s_t~|~ s_t,a_t,\Dset],    
\end{equation}
for all $s_t,a_t,z,\hat{s}_{t+1}$ with $\prob[s_t,a_t,z|q,\Dset]>0$ and define a Markov policy $\pi$ as
\begin{equation}
    \pi(\hat{a}|s_t,z) = \prob[\hat{a} = a_t|s_t,z,\Dset].
\end{equation}
Then for any $z$ with $\prob[z|q,\Dset]>0$ and any $\tau$,
\begin{equation}
    \prob[\tau~|~\pi_z, \mdp] > 0 ~~\mathrm{if~and~only~if}~~ \prob[\tau~|~z,\Dset] > 0.
\end{equation}
}

\begin{figure}
    \centering
    \includegraphics[width=.3\textwidth]{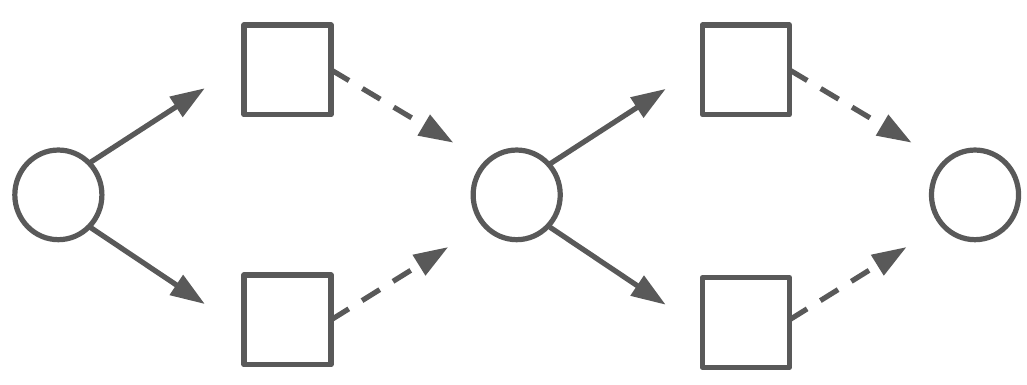}
    \caption{Deterministic environment used in the counter-example described in Appendix~\ref{sec:bad-consistency}. 
    Circles represent states and squares represent actions; solid arrows represent choice of actions and dashed arrows represent environment dynamics.}
    \label{fig:counter-example}
\end{figure}

\paragraph{Counter-example.} A simple counter-example may be constructed by considering the Markovian environment displayed in Figure~\ref{fig:counter-example}. The environment has three states. The first state gives a choice of two actions ($a_0\in\{0,1\}$), and each action deterministically transitions to the same second state. 
The second state again provides a choice of two actions ($a_1\in\{0,1\}$), and each of these again deterministically transitions to the same terminal state. Thus, episodes in this environment are uniquely determined by choice of $a_0,a_1$. There are four unique episodes:
\begin{align}
    \tau_0 &= \langle a_0=0,a_1=0\rangle, \\
    \tau_1 &= \langle a_0=1,a_1=1\rangle, \\
    \tau_2 &= \langle a_0=0,a_1=1\rangle, \\
    \tau_3 &= \langle a_0=1,a_1=0\rangle.
\end{align}
We now construct $q$ as a deterministic function, clustering these four trajectories into two distinct $z$:
\begin{align}
    z_0 &= q(\tau_0) = q(\tau_1), \\
    z_1 &= q(\tau_2) = q(\tau_3).
\end{align}
Suppose $\Dset$ includes $\tau_0,\tau_1,\tau_2,\tau_3$ with equal probability. Since the environment is deterministic, the conditions of Theorem 2.1 in~\citet{paster2022you} are trivially satisfied.
Learning a policy $\pi$ with respect to $z_0$ yields
\begin{align}
    \pi(\cdot | s_0, z_0) &= [0.5, 0.5], \\
    \pi(\cdot | s_1, z_0) &= [0.5, 0.5].
\end{align}
However, it is clear that interacting with $\pi(\cdot|\cdot,z_0)$ in the environment will lead to $\tau_2,\tau_3$ with non-zero probability, while $\tau_2,\tau_3$ are never associated with $z_0$ in the data $\Dset$. Contradiction.

\section{Pseudocode for \method training}\label{app:pseudo}

\begin{algorithm}[h]
\caption{Training with Dichotomy of Control}
\begin{algorithmic}
\STATE \textbf{Inputs} Offline dataset $\Dset=\{\tau^{(m)}\}_{m=1}^M$ where $\tau^{(m)}=(s_t^{(m)}, a_t^{(m)}, r_t^{(m)})_{t=0}^H$ with initial states $\{s_0^{(m)}\}_{m=1}^M$ and initial return-to-go values $\{R^{(m)}\}_{m=1}^M$, a parametrized distribution $q_\phi(\cdot)$, a policy $\pi_{\theta_1}(\cdot, \cdot)$, a value function $V_{\theta_2}(\cdot)$, a prior $p_\psi(\cdot)$, an energy function $f_w(\cdot)$, a fixed distribution $\rho(r, s')$, learning rates $\eta$, and training batch size $B$.
\WHILE{training has not converged}
   \STATE Sample batch $\{(\tau=(s_t, a_t, r_t)_{t=0}^H)^{(m)}\}_{m=1}^B$ from $\Dset$, for $m=1,\dots,B$.
   \STATE Sample $z$ from $q_\phi(\tau)$ with reparametrization.
   \STATE Compute $\DOC+\mathcal{L}_{aux}$ according to Equation~\ref{eq:doc} and Equation~\ref{eq:doc-aux}.
   \STATE Update $\phi\leftarrow \phi - \eta \nabla_{\phi}\hat{\mathcal{L}}$,  $\psi\leftarrow \psi - \eta \nabla_{\psi}\text{stopgrad}(\hat{\mathcal{L}}, \phi)$, $w\leftarrow w + \eta \nabla_{w}\hat{\mathcal{L}}$, $\theta_1\leftarrow \theta_1 - \eta \nabla_{\theta_1}\hat{\mathcal{L}}$, $\theta_2\leftarrow \theta_2 - \eta \nabla_{\theta_1}\hat{\mathcal{L}}$.
\ENDWHILE
\textbf{return}\,\,$\pi_{\theta_1}(\cdot,\cdot),V_{\theta_2}(\cdot),p_\psi(\cdot)$
\end{algorithmic}
\label{algo:doc-train}
\end{algorithm}

\clearpage
\newpage
\section{Experiment Details}\label{app:exp_details}
\subsection{Hyperparameters.}
We use the same hyperparameters as the publically available Decision Transformer~\citep{chen2021decision} implementation. For VAE, we additionally learn a future and a prior both parametrized the same as the policy using transformers with context length 20. All models are trained on NVIDIA GPU P100.
\begin{table}[ht]
\caption{Hyperparameters of Decision Transformer, future-conditioned VAE, and Dichotomy of Control.}
\vskip 0.15in
\begin{center}
\begin{small}
\begin{tabular}{ll}
\toprule
\textbf{Hyperparameter} & \textbf{Value}  \\
\midrule
Number of layers & $3$  \\
Number of attention heads    & $1$  \\
Embedding dimension    & $128$  \\
Latent future dimension    & $128$  \\
Nonlinearity function & ReLU \\
Batch size   & $64$ \\
Context length $K$ & $20$ FrozenLake, HalfCheetah, Hopper, Humanoid, AntMaze \\
               & $5$ Reacher \\
Future length $K_f$ & Same as context length $K$ \\               
Return-to-go conditioning for DT  & $1$ FrozenLake \\
                            & $6000$ HalfCheetah \\
                            & $3600$ Hopper \\
                            & $5000$ Humanoid \\
                            & $50$ Reacher \\
                            & $1$ AntMaze \\
Dropout & $0.1$ \\
Learning rate & $10^{-4}$ \\
Grad norm clip & $0.25$ \\
Weight decay & $10^{-4}$ \\
Learning rate decay & Linear warmup for first $10^5$ training steps \\
$\beta$ coefficient & $1.0$ for DoC, Best of $0.1, 1.0, 10$ for VAE \\
\bottomrule
\end{tabular}
\end{small}
\label{tab:hyperparameters}
\end{center}
\end{table}

\subsection{Details of the offline datasets}
\paragraph{FrozenLake.} We train a DQN~\citep{mnih2013playing} policy for 100k steps in the original 4x4 FrozenLake Gym environment with stochasticity level $p=\frac{1}{3}$. We then modify $p$ to simulate environments of different stochasticity levels, while collecting 100 trajectories of maximum length 100 at each level using the trained DQN agent with probability $\epsilon$ of selecting a random action as opposed to the action output by the DQN agent to emulate offline data with different quality.

\paragraph{Gym MuJoCo.} We train SAC~\citep{haarnoja2018soft} policies on the original set of Gym MuJoCo environments for 100M steps. To simulate stochasticity in these environments, we modify the original Gym MuJoCo environments by introducing noise to the actions before inputting the action to the physics simulator to compute rewards and next states. The noise has 0 mean and standard deviation of the form $(1 - e^{-0.01 \cdot t}) \cdot \sin(t) \cdot \sigma$ where $t$ is the step number and $\sigma\in[0,1]$. We then collect 1000 trajectories of 1000 steps each for all environments except for Reacher (which has 50 steps in each trajectory) in the stochastic version of the environment using the SAC policy to acquire the offline dataset for training.

\paragraph{AntMaze.} For the AntMaze task, we use the AntMaze dataset from D4RL~\citep{fu2020d4rl}, which contains 1000 trajectories of 1000 steps each. We add gaussian noise with standard deviation 0.1 to the rewards in the dataset uniformly with probability 0.1 to both the offline dataset and during environment evaluation to simulate stochastic rewards from the environment.

\clearpage
\newpage
\section{Additional Results}\label{app:exp_results}
\subsection{FrozenLake with different offline dataset quality.}
\begin{figure}[h!]
    \centering
    \includegraphics[width=\linewidth]{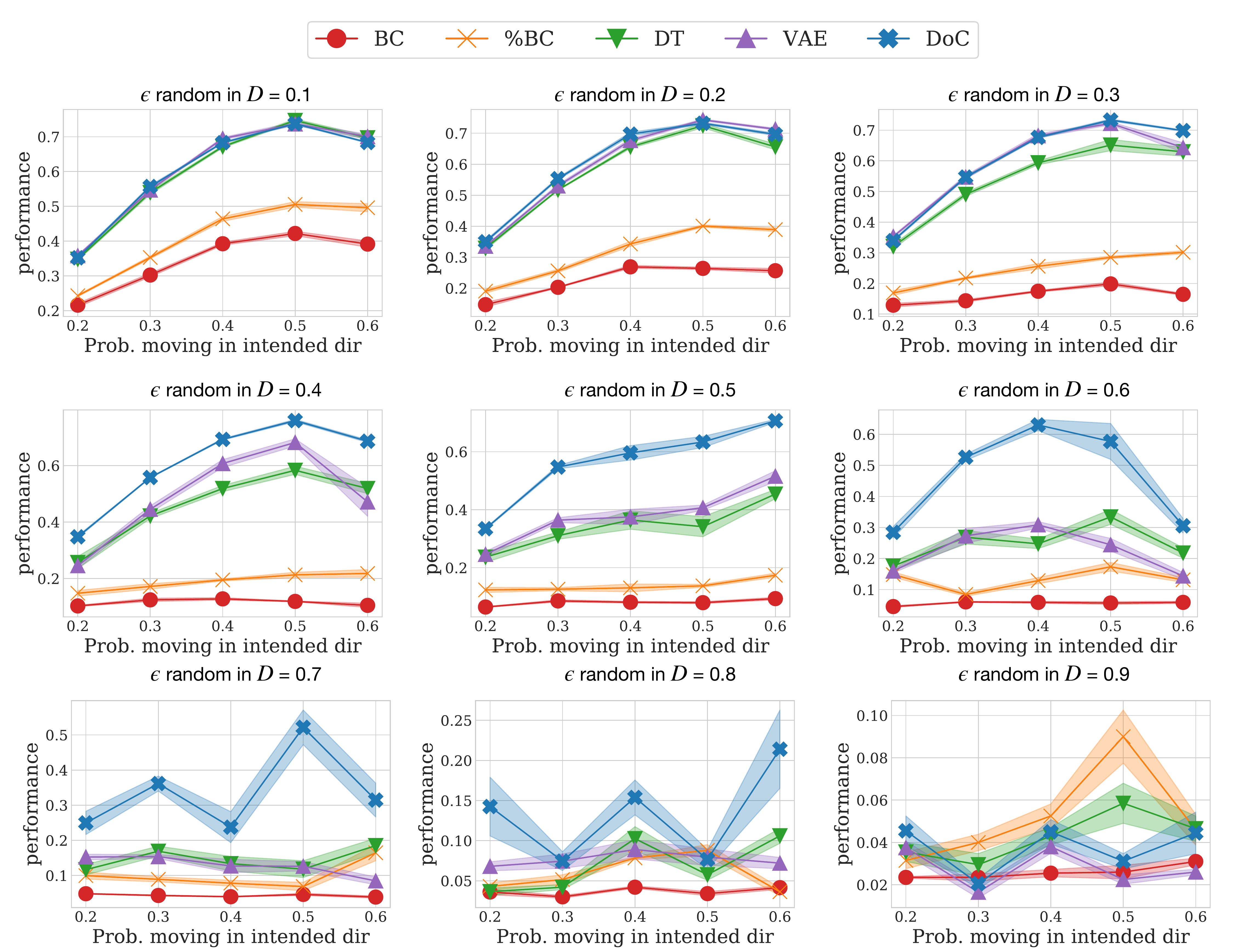}
    \caption{Average performance (across 5 seeds) of \method and baselines on FrozenLake with different levels of stochasticity ($p$) and offline dataset quality ($\epsilon$). \method outperforms DT and future VAE with bigger gains the offline data is less optimal.}
    \label{fig:lake_app}
\end{figure}

\end{document}